\documentclass[11pt]{article}
\PassOptionsToPackage{table}{xcolor}
\usepackage[final]{acl}

\usepackage{times}
\usepackage{latexsym}
\usepackage[T1]{fontenc}
\usepackage[utf8]{inputenc}
\usepackage{microtype}
\usepackage{graphicx}
\usepackage{booktabs}
\usepackage{amsmath}
\usepackage{xcolor}
\usepackage{subcaption}
\usepackage{verbatim}
\usepackage{placeins}
\usepackage{float}
\usepackage{flafter}

\makeatletter
\newcounter{savedlinenumber}
\newcommand{\suspendLineNumbersIfReview}{%
  \ifacl@linenumbers
    \global\setcounter{savedlinenumber}{\value{linenumber}}%
    \nolinenumbers
  \fi}
\newcommand{\resumeLineNumbersIfReview}{%
  \ifacl@linenumbers
    \linenumbers
    \global\setcounter{linenumber}{\value{savedlinenumber}}%
  \fi}
\makeatother

\AtBeginEnvironment{figure}{\suspendLineNumbersIfReview}
\AtEndEnvironment{figure}{\resumeLineNumbersIfReview}
\AtBeginEnvironment{figure*}{\suspendLineNumbersIfReview}
\AtEndEnvironment{figure*}{\resumeLineNumbersIfReview}
\AtBeginEnvironment{table}{\suspendLineNumbersIfReview}
\AtEndEnvironment{table}{\resumeLineNumbersIfReview}
\AtBeginEnvironment{table*}{\suspendLineNumbersIfReview}
\AtEndEnvironment{table*}{\resumeLineNumbersIfReview}

\definecolor{darkblue}{rgb}{0, 0, 0.5}
\hypersetup{colorlinks=true, citecolor=darkblue, linkcolor=darkblue, urlcolor=darkblue}
\providecommand{\accnum}[1]{\makebox[2.65em][r]{#1}}
\providecommand{\accbox}[2]{\begingroup\setlength{\fboxsep}{0.4pt}\colorbox{#1}{\accnum{#2}}\endgroup}

\title{Mechanistic Diagnostics of Spatial Lexical Bias \\
in Multimodal Large Language Model Spatial Reasoning}

\author{
  \textbf{Chuang Ma\textsuperscript{1,2}} \quad
  \textbf{Qianying Liu\textsuperscript{2,$\dagger$}} \quad
  \textbf{Tomoyuki Obuchi\textsuperscript{1,3}} \quad
  \textbf{Fei Cheng\textsuperscript{1}} \\
  \textbf{Wang Yang\textsuperscript{4}} \quad
  \textbf{Sudong Cai\textsuperscript{5}} \quad
  \textbf{Shuyuan Zheng\textsuperscript{6}} \quad
  \textbf{Akiko Aizawa\textsuperscript{7,2}} \quad
  \textbf{Sadao Kurohashi\textsuperscript{2}} \\[3pt]
  \textsuperscript{1}Kyoto University \quad
  \textsuperscript{2}NII LLMC \quad
  \textsuperscript{3}RIKEN AIP \quad
  \textsuperscript{4}Case Western Reserve University \\
  \textsuperscript{5}The Hong Kong Polytechnic University \quad
  \textsuperscript{6}The University of Osaka \quad
  \textsuperscript{7}University of Tokyo \\[3pt]
  \texttt{\{ying, aizawa, kurohashi\}@nii.ac.jp \quad zheng@ist.osaka-u.ac.jp} \\
  \texttt{wxy320@case.edu \quad sudong.cai@polyu.edu.hk} \\
  \texttt{\{ma.chuang.52h@st, obuchi@i, feicheng@i\}.kyoto-u.ac.jp}
}

\begin{document}
\maketitle

\begingroup
\renewcommand{\thefootnote}{$\dagger$}
\footnotetext{Corresponding author.}
\endgroup

\begin{abstract}
Multimodal large language models (MLLMs) remain unreliable on spatial
multiple-choice questions, and their failures are often attributed to poorly
attended visual information. We identify a complementary failure mode,
spatial lexical bias: a spatial relation word added to the answer
options can act as a lexical-semantic distractor that draws the model's
decision toward that option.
Using nine open-weight MLLMs, we show that this phenomenon is widespread.
We then isolate diagnostic cases in which a model answers a binary spatial
question correctly yet consistently chooses a newly added third spatial
option, which we call binary-stable but ternary-fragile cases. Leveraging mechanistic interpretability tools on these cases, we find that
the failure arises on the language side rather than the visual side: visual attention analyses and residual-stream probes
show the correct spatial relation remains internally available, while irrelevant-option controls, activation patching, and sparse
component interventions trace the bias to specific LLM-side channels and
neurons.
Accordingly, we show that a lightweight LLM-only DPO update on tiny
single-object-pair synthetic data mitigates the bias, lifting four-way robust
accuracy by up to 100 points on synthetic data, and by 68.0, 32.6, and 20.1
points on broader evaluation datasets WhatsUp, SpatialMQA-Direct, and VSR.
\end{abstract}

\section{Introduction}
\label{sec:intro}

\begin{figure}[t!]
\centering
\includegraphics[width=\columnwidth]{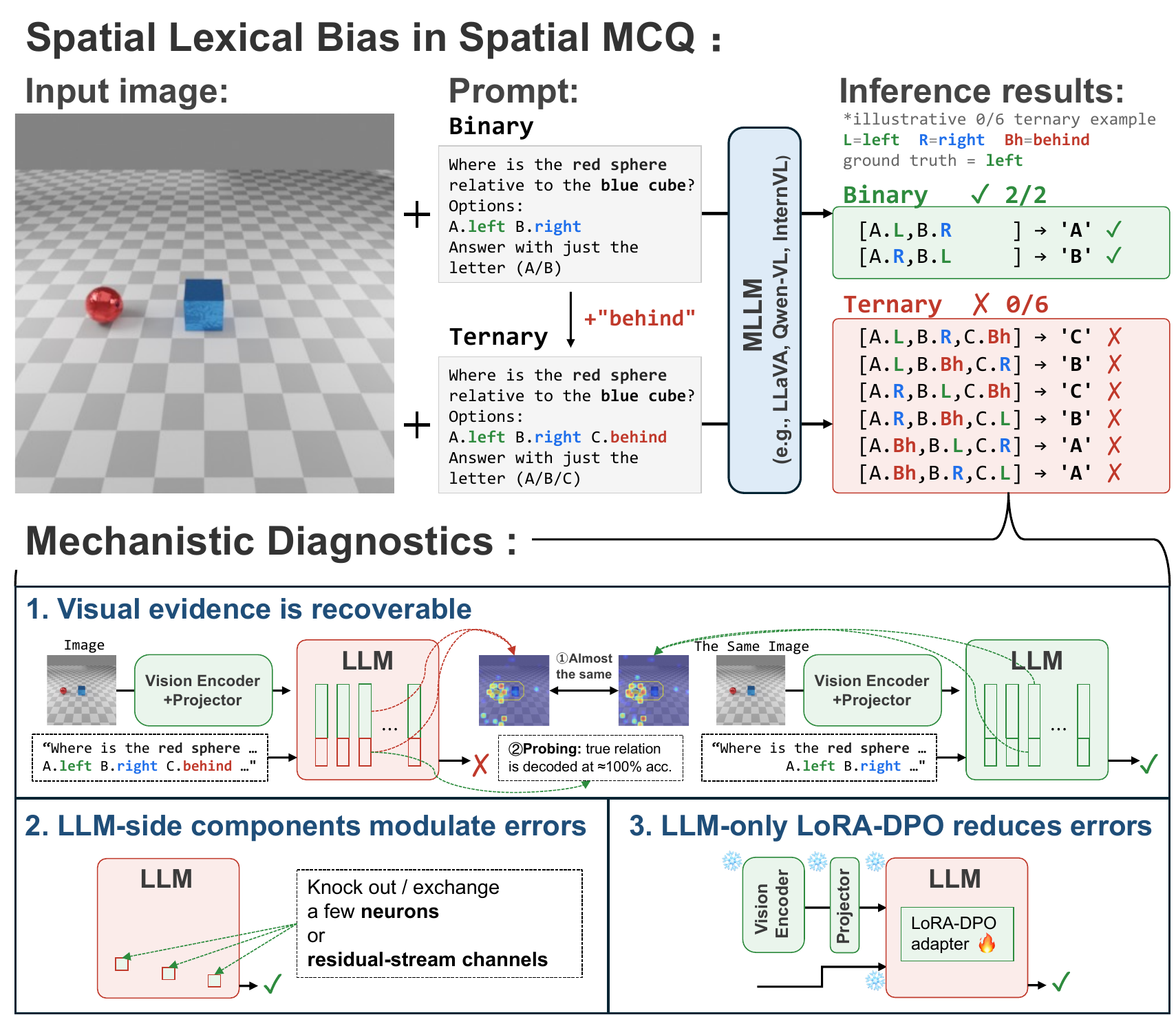}
\caption{\textbf{Failure pattern and evidence chain.}
\textbf{(Top)} One illustrative sample answers the binary \{Left, Right\}
question correctly under both option orders, but chooses the added ``Behind''
option in all six ternary orderings. 
\textbf{(Bottom)} Our diagnostics test whether the relation remains
recoverable from last-token residual-stream vectors, identify LLM-side
residual-stream channels and MLP neurons whose intervention changes selected
failures, and evaluate an LLM-only LoRA-DPO repair.}
\label{fig:overview}
\end{figure}

Spatial reasoning evaluates whether multimodal large language models
ground language in visual spatial structure, supporting applications that require models to understand object locations~\citep{liu-etal-2023-visual,kamath-etal-2023-whats,10.1007/978-3-031-73337-6_9,liu-etal-2025-multimodal-large,Chen_2024_CVPR}.
Recent studies have repeatedly shown that MLLMs remain unreliable on spatial reasoning tasks, and prior work often attributes these failures to weak visual grounding, missing visual evidence, or incorrect visual attention~\citep{Tong_2024_CVPR,Leng_2024_CVPR,Huang_2024_CVPR,pmlr-v267-chen25cr}. In this study, we begin from a simple but revealing phenomenon that challenges such purely visual-information-loss explanation. As illustrated in Figure~\ref{fig:overview}, given the same image and the same question, a model can answer correctly when choosing between the binary options \textit{left} and \textit{right}. Yet when a new spatial option, \textit{behind}, is added, the model can collapse and consistently choose the newly introduced option, regardless of answer order. For such specific option combinations, which we refer to as \textit{binary-stable but ternary-fragile} (BSTF) cases, this collapse is systematic at the case level: a model can solve the binary classification nearly perfectly, yet once the particular spatial alternative is introduced, it fails across the ternary samples and consistently prefers the new option. This collapse raises a diagnostic question: \textbf{when the model fails under the larger option set, has it truly lost the visual relation, or has the added spatial word changed the language-side decision process?}

We investigate this question in a controlled synthetic setting (Figure~\ref{fig:overview}). If the added option disrupted visual grounding, we would expect weaker access to the relevant visual evidence. We observe the opposite. Attention maps show that binary-correct and ternary-wrong prompts attend to similar object regions, so the added option does not redirect the model away from the relevant visual evidence. Layer-wise probing of the last-token residual stream further recovers the correct spatial relation almost perfectly, even when the model selects the newly added wrong option.
These results indicate that the collapse is not primarily caused by visual information loss: the correct relation is still present in the model's internal representation, while the final answer is redirected by the changed option set.

We therefore hypothesize the failure instead points to a language-side \textbf{spatial lexical bias} among spatial answer options. We argue that spatial answer words are not neutral labels: words such as left, right, front, and behind carry structured lexical-semantic relations inside the LLM, and a newly added spatial word can act as a lexical-semantic distractor that interferes with the LLM-side decision process. We compare spatial distractors with irrelevant options such as \textit{violin}. Irrelevant options cause only modest degradation, whereas specific spatial alternatives can collapse the prediction and are consistently selected as the wrong answer. This contrast suggests that the failure is not merely caused by increasing the number of choices, but by the semantic structure among spatial words. We further show that the same distractors extend from ternary to four-way settings, indicating that the model has stable preferences over particular spatial options. 

To move beyond behavioral observation, we use interpretability tools to locate where the lexical-semantic attraction emerges. Activation patching shows that exchanging residual-stream states between binary-correct and ternary-wrong prompts can rescue or corrupt the answer, with the transition emerging around the middle-to-late layers. Sparse interventions further identify residual-stream channels and MLP neurons at the added-option position whose knockout partially restores correct predictions. These results provide causal evidence that specific LLM-side layers and components modulate the spatial-option bias.

Finally, we test whether spatial lexical bias can be mitigated through a lightweight LLM-side update. Using a minimal synthetic preference dataset built from a single object-pair setting, we apply Direct Preference Optimization (DPO; \citealp{NEURIPS2023_a85b405e}) only to the LLM decoder, with the vision encoder and projector kept frozen. Despite the simplicity of the training source, this LLM-only debiasing substantially improves spatial reasoning under four-way option sets, with gains of up to 100 percentage points on our controlled synthetic data. The improvement also transfers to broader and noisier public benchmarks, reaching gains of up to 68.0 points on WhatsUp, 32.6 points on SpatialMQA-Direct, and 20.1 points on VSR. This suggests that the bias is not limited to the specific binary-to-ternary cases but reflects a broader source of spatial reasoning failure, repairable by adjusting the language-side decision process. The repair also extends to relations absent from training and leaves general capabilities intact. DPO is not the only effective repair, as we also confirm that SFT achieves similar improvements.

In summary, our contributions are threefold:
\begin{itemize}
\item We identify \textit{binary-stable but ternary-fragile} cases, a systematic failure pattern in which MLLMs solve a binary spatial contrast but collapse after a particular spatial alternative is added. Attention and probing analyses show the correct visual relation remains internally recoverable, indicating the failure is not simply caused by visual information loss.
\item We introduce \textit{spatial lexical bias} as an LLM-side mechanism behind these failures. Irrelevant-option controls and ternary-to-four-way extensions show that spatial answer words interact in a structured way, while activation patching and sparse component interventions localize the bias to manipulable LLM-side states and components.
\item We show that this bias can be reduced with LLM-only DPO debiasing. A minimal single-object-pair synthetic preference dataset substantially improves spatial robustness and transfers to broader, noisier evaluation data and to unseen relations while preserving general capabilities.
\end{itemize}

\section{Related Work}
\label{sec:related}

We situate this work along three lines: spatial reasoning in MLLMs, MCQ position
bias, and mechanistic interpretability.

\paragraph{Spatial reasoning in MLLMs.}
Spatial benchmarks test whether vision-language models ground relation words in
visual structure. This line runs from early spatial-relation datasets
\citep{Yang_2019_ICCV,NEURIPS2020_76dc611d} to recent MLLM spatial-reasoning
benchmarks and systems
\citep{kamath-etal-2023-whats,liu-etal-2023-visual,liu-etal-2025-multimodal-large,NEURIPS2024_f38cb4cf,jia2026omnispatial}. Many analyses explain failures through
weak visual grounding, underused visual information, or misplaced visual focus
\citep{Tong_2024_CVPR,10.1007/978-981-96-0917-8_17,pmlr-v267-chen25cr}. Our study is
complementary: we isolate cases where the visual contrast is solved in binary
form, but fails after a plausible spatial option is added.

\paragraph{MCQ position bias.}
MCQ evaluation can be sensitive to answer letters and option order in LLMs
\citep{robinson2023leveraging,ICLR2024_54dd9e0c,pezeshkpour-hruschka-2024-large,wang2024look}
and LVLMs \citep{atabuzzaman-etal-2025-benchmarking}. We therefore use complete option-order
permutations and define a sample as solved only when all orderings are correct.
This makes the BSTF diagnostic conservative with respect to position bias
(Appendix~\ref{app:position_bias_breakdown}).

\paragraph{Mechanistic interpretability.}
Mechanistic interpretability asks whether information is represented and whether
intervening on it changes behavior. We use standard tools for this purpose:
probing for recoverability
\citep{hewitt-liang-2019-designing,belinkov-2022-probing}, and activation patching or
sparse intervention for causal leverage
\citep{NEURIPS2022_6f1d43d5,dai-etal-2022-knowledge,wang2023interpretability,NEURIPS2023_34e1dbe9,ICLR2024_06a52a54,ameisen2025circuit}. We use them in a
scoped way: not to claim a complete circuit, but to test whether the visual
relation remains available and whether LLM-side states can modulate the failure.

\section{BSTF Cases as a Diagnostic Tool}
\label{sec:setting}

This section builds the diagnostic tool used later. First, we make a clean
setting: synthetic scenes, four directions, and complete option-order
evaluation. Second, strict scoring exposes sharp drops once a spatial option is
added. Third, we select the clearest \emph{binary-stable but
ternary-fragile} (BSTF) cases: binary solved, one added spatial option breaks
the answer.

\paragraph{Task and prompts.}
The task is deliberately small so the failure is easy to interpret. Each
question asks for one of four directions:
\{\emph{left}, \emph{right}, \emph{front}, \emph{behind}\}. We test binary,
ternary, and four-way option sets under complete answer-order permutations. A
later control replaces the added spatial option with an irrelevant option
(Section~\ref{sec:irrelevant}); full templates are in
Appendix~\ref{app:prompt_configurations}.

\paragraph{Controlled synthetic datasets.}
The synthetic datasets keep the diagnostic setting clean. With visual and
annotation noise minimized, BSTF-case mining can focus on whether answer-option
words disrupt four-direction judgments. We render three balanced controlled
synthetic sets,
Sphere--Cube Raw (\textbf{CS}),
Sphere--Dog Raw (\textbf{DR}), and Sphere--Dog Outdoor (\textbf{DO}); each
contains 2{,}000 images with 500 per relation
(Figure~\ref{fig:dataset_examples_main}). Rendering details are in
Appendix~\ref{app:synthetic_datasets}. Three public benchmarks
are used only as out-of-distribution probes in Section~\ref{sec:repair}.

\paragraph{MLLMs.}
The model sweep checks whether the pattern is tied to one architecture family.
We evaluate 9 open-weight MLLMs from LLaVA, Qwen, and InternVL families,
covering 1B--8B parameter scales (Table~\ref{tab:models_overview}).

\paragraph{Evaluation metrics.}
The metric controls MCQ position bias. For every option set, we permute option
text across fixed A/B/C/D slots (Figure~\ref{fig:overview}). Our main metric is
\emph{sample-wise robust accuracy}: a sample is correct only if all orderings
are correct. Thus binary, ternary, and four-way prompts require all 2, 6, and 24
orderings to be correct. Appendix~\ref{app:inference_settings} details the
model-loading, decoding, and answer-extraction procedures.

\begin{figure}[t]
\centering
\setlength{\abovecaptionskip}{2pt}
\includegraphics[width=\columnwidth]{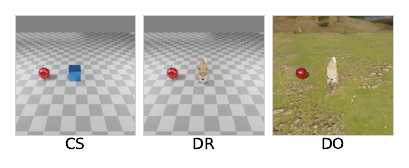}
\caption{\textbf{Controlled synthetic examples.}}
\label{fig:dataset_examples_main}

\vspace{3pt}
\begingroup
\scriptsize
\setlength{\tabcolsep}{4pt}
\renewcommand{\arraystretch}{1.04}
\setlength{\abovecaptionskip}{3pt}
\begin{tabular}{@{}llccc|c@{}}
\toprule
\textbf{Model} & \textbf{Prompt} & \textbf{CS} & \textbf{DR} & \textbf{DO} & \textbf{Avg.} \\
\midrule
\rowcolor{gray!8}LLaVA & Binary & \accbox{red!8}{21.3} & \accbox{red!8}{35.8} & \accbox{gray!10}{54.6} & \accbox{red!8}{37.2} \\
\rowcolor{gray!8}1.5-7B & Ternary & \accbox{red!8}{0.3}  & \accbox{red!8}{1.6}  & \accbox{red!8}{23.0} & \accbox{red!8}{8.3} \\
\rowcolor{gray!8}        & Four-way & \accbox{red!8}{0.0}  & \accbox{red!8}{0.0}  & \accbox{red!8}{2.8} & \accbox{red!8}{0.9} \\
\addlinespace[1pt]
LLaVA & Binary & \accbox{blue!18}{72.3} & \accbox{blue!26}{83.1} & \accbox{blue!26}{83.8} & \accbox{blue!18}{79.7} \\
v1.6-Vicuna & Ternary & \accbox{gray!10}{58.1} & \accbox{blue!11}{65.5} & \accbox{gray!10}{58.0} & \accbox{blue!11}{60.5} \\
            & Four-way & \accbox{gray!10}{50.0} & \accbox{red!8}{44.8}   & \accbox{red!8}{27.9} & \accbox{red!8}{40.9} \\
\addlinespace[1pt]
\rowcolor{gray!8}LLaVA & Binary & \accbox{gray!10}{47.4} & \accbox{red!8}{42.0} & \accbox{gray!10}{54.2} & \accbox{gray!10}{47.9} \\
\rowcolor{gray!8}v1.6-Mistral & Ternary & \accbox{red!8}{16.7}   & \accbox{red!8}{19.1} & \accbox{red!8}{32.7} & \accbox{red!8}{22.8} \\
\rowcolor{gray!8}             & Four-way & \accbox{red!8}{0.0}    & \accbox{red!8}{0.0}  & \accbox{red!8}{2.6} & \accbox{red!8}{0.9} \\
\bottomrule
\end{tabular}
\captionof{table}{\textbf{Illustrative spatial MCQ weakness under strict scoring.}
Sample-wise robust accuracy (\%) on controlled synthetic sets before BSTF-case
selection.}
\label{tab:pre_mining_synthetic}
\endgroup
\end{figure}

\begin{figure}[t]
\centering
\includegraphics[width=\columnwidth]{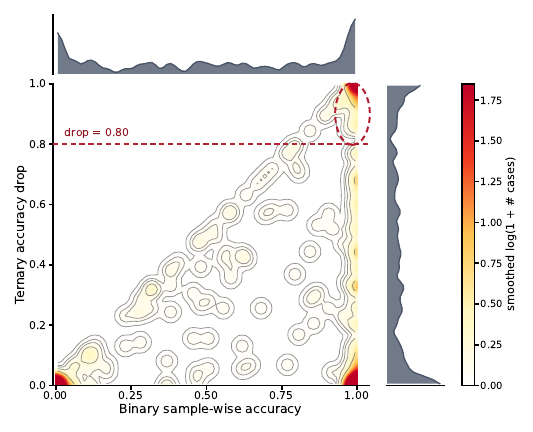}
\caption{\textbf{BSTF diagnostic-case selection.}
Smoothed density of binary vs.\ ternary-drop sample-wise robust accuracy over
all cases in the synthetic-dataset $\times$ 9-MLLM sweep. The dashed line
marks the 80~pp drop threshold; the dashed ellipse marks the selected BSTF
region (55 cases, 7 models, three families).}
\label{fig:highhigh_scatter}
\end{figure}

The strict metric shows why a diagnostic is needed. Adding spatial options leaves LLaVA-family spatial MCQ performance far from
solved (Table~\ref{tab:pre_mining_synthetic}), raising the question of what
causes these option-set failures.

\subsection{Binary-Stable but Ternary-Fragile (BSTF) Cases}
\label{sec:highhigh}

We now turn the broad failure pattern into a concrete diagnostic tool. A BSTF
case is a fixed combination of model, dataset, true relation, binary competitor,
and added spatial alternative. We keep only the clearest cases: binary robust
accuracy must be perfect (1.00), and adding one spatial alternative must drop
ternary accuracy by at least 80 percentage points
(Figure~\ref{fig:highhigh_scatter}). This high bar yields 55 cases across
7 models spanning all three architecture families
(LLaVA-1.5-7B, LLaVA-v1.6-Vicuna-7B, LLaVA-v1.6-Mistral-7B, InternVL2-1B,
InternVL2.5-1B, Qwen2-VL-2B, Qwen3-VL-8B). The two remaining models still show
sizable but sub-threshold maximum drops (57.0 and 76.8~pp for Qwen2-VL-7B and
Qwen3-VL-2B).

\textbf{BSTF cases are extreme exemplars, not exhaustive.} The
binary-stable / ternary-fragile drop pattern is not confined to these 55 cases
and appears broadly across our permutation sweep. We use the strict 80~pp set
only as a low-noise tool for the mechanism analyses in
Sections~\ref{sec:mechanism}--\ref{sec:repair}; the repair study later returns
to all 9 models and three additional public datasets
(Section~\ref{sec:repair}). Importantly, BSTF cases serve only as
clean, controlled diagnostic examples and do not imply that MLLM spatial
accuracy decreases as the number of answer options increases; on BSTF cases, replacing the incorrect binary option with the added spatial
option, still with two options, drops strict accuracy from 1.000 to 0.163
(Appendix~\ref{app:three_to_four_extension}).

\textbf{Binary successes on these cases are not near ties.} On the
matched binary prompts the mean probability of the correct option is 0.828;
it falls to 0.297 once the added option is present, decreasing in all 55
cases (Appendix~\ref{app:confidence}).

\section{Visual Evidence Remains Recoverable}
\label{sec:spatial_evidence_present}

This section asks whether BSTF failures are caused by losing visual evidence.
We test this in two simple steps. First, attention visualization asks whether
the ternary-error prompt still looks at the relevant objects. Second, last-token
residual-stream probes ask whether the true relation is still decodable across
layers. If both tests remain positive, the added spatial option has not erased
the visual relation.

\begin{figure}[t]
\centering
\includegraphics[width=\columnwidth]{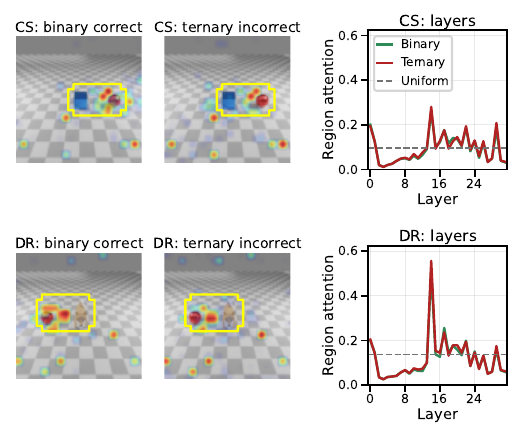}
\caption{\textbf{Diagnostic last-token visual attention.}
Rows show two representative selected BSTF cases for LLaVA-1.5-7B, one from CS
and one from DR. Yellow contours mark the bi-object region of interest (ROI).
Columns show binary-correct heatmaps, ternary-incorrect heatmaps, and filtered
per-layer ROI-attention curves. Heatmaps are from layer~14 at the last token.
The rightmost column shows the fraction of attention inside the ROI
after the case mean is projected out, computed over binary-correct and
ternary-incorrect prompt records.}
\label{fig:diagnostic_lasttoken_attn}
\end{figure}

\begin{figure}[t]
\centering
\includegraphics[width=\columnwidth]{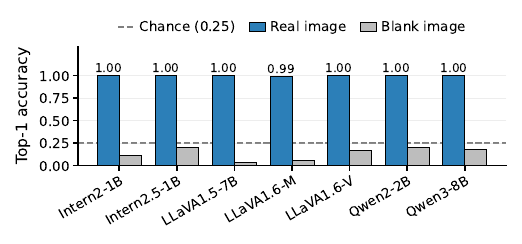}
\caption{\textbf{Probing diagnostic.}
A four-way linear probe reads out the true relation from added-spatial-option
error rows.
Real-image probes are near-perfect for every model (macro accuracy \(0.998\));
blank-image controls collapse below chance (macro accuracy \(0.137\)). The dashed
line marks four-way chance (\(0.25\)).}
\label{fig:probe_per_model}
\end{figure}

\subsection{Attention Visualization Shows Relevant Objects Are Still Tracked}
\label{sec:last_token_attention_visualization}

We first use attention visualization to test whether the added spatial option
changes where the model looks. If the added option damages visual grounding, the
ternary-error prompt should shift away from the relevant objects. For
visualization, we select two representative BSTF cases for LLaVA-1.5-7B, one
from CS and one from DR. We compare last-token attention to image patches between
binary-correct and ternary-incorrect prompts. The region of interest (ROI) is the
patch-level enclosing region of the two objects. For visual clarity, we project out
the per-family case-mean attention direction and clip negative values. This setup
directly tests whether the added spatial option changes the visual region the
model uses.

The attention visualization gives a direct answer: the visual focus does not
move. Figure~\ref{fig:diagnostic_lasttoken_attn} shows two selected BSTF cases:
CS and DR. Both are maximally fragile under our strict metric: binary accuracy
is \(1.00\), while ternary accuracy is \(0.00\). Yet the binary-correct and
ternary-incorrect heatmaps are nearly the same: in both cases they focus on the
same bi-object region. Their layer-14 ROI masses are also similar:
\(0.26/0.28\) for CS and \(0.51/0.56\) for DR, above uniform ROI baselines
(\(0.096\) and \(0.137\)). Thus, at the attention level, the added spatial
distractor does not visibly disrupt access to the relevant visual region.

This pattern is not specific to the two illustrated cases. All 55 BSTF cases show the same behavior, with a mean scene-level Pearson
correlation of 0.978 between the binary and ternary attention maps. Details
are given in Appendix~\ref{app:attention_allcases}.

\subsection{Last-Token Probes Recover the True Relation Across Layers}
\label{sec:spatial_probe}

Attention provides a visual diagnostic, but we next test the representation more
directly: whether the true relation is still stored inside the last-token
residual stream. For each model, we train a four-way linear probe on
scene-disjoint correct rows, then evaluate it on added-spatial-option error
rows. This is a retained-information diagnostic, not a causal-use test: it asks
whether the true relation is still linearly present. Appendix~\ref{app:probing_details}
gives the full design and controls.
This probe directly tests whether the correct visual relation is still
recoverable after the model has made the wrong ternary choice.

The probe result gives the stronger answer: the true relation is still
recoverable. Across all seven instrumented models, the probe recovers the true
relation on added-spatial-option error rows with macro accuracy \(0.998\). This
is not only a last-layer effect: the full layer analysis in
Appendix~\ref{app:probing_details} shows that recoverability stays near ceiling
across almost the whole stack, and already exceeds \(0.92\) at the first layer.
As a control, we keep the prompt fixed but replace the image with a blank
canvas; macro accuracy drops to \(0.137\), below chance.
A randomized-target control likewise rules out probe memorization, as
probes retrained on permuted labels collapse to chance on held-out scenes
(Appendix~\ref{app:probe_selectivity}).
Thus, even in the most
extreme selected failures, the correct
visual relation remains stored in the last-token residual stream rather than
being erased by the added spatial option.

\begin{figure}[t]
\centering
\includegraphics[width=\columnwidth]{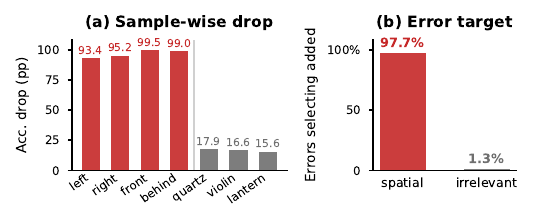}
\caption{\textbf{Irrelevant third-option control.}
\textbf{(a)} Sample-wise robust drops for the original spatial added options
and the three irrelevant words. \textbf{(b)} Among prompt-level errors,
spatial-third-option errors almost always select the newly added option,
whereas irrelevant-option errors rarely select the irrelevant word itself.}
\label{fig:irrelevant_control}
\end{figure}

Together, the two diagnostics leave a simple conclusion: the model still looks
at the relevant objects, and the true relation is still decodable. The failure
therefore is not primarily visual evidence loss. We next localize the downstream
decision problem inside the LLM-side computation.

\section{LLM-Side Diagnostics of Spatial Lexical Bias}
\label{sec:mechanism}
\label{sec:llm_localization}

We now test the LLM-side explanation: the added spatial word interacts with
other spatial answer words through lexical-semantic structure. The localization
logic has three steps. First, we test whether the effect exists as a
spatial-word-specific bias, rather than a generic cost of adding another option
or a position-bias artifact. Second, we ask at what depth an LLM state becomes
sufficient to switch the answer. Third, we ask whether local components at the
added-option token can modulate the same error. Thus this section moves from
\emph{existence and specificity}, to \emph{layer-level localization}, to
\emph{local component localization}.

\subsection{Irrelevant Options Reveal Spatial Semantic Structure}
\label{sec:irrelevant}

The first step is the existence and specificity test: does the failure require
a spatial word, or would any added option cause the same drop? For each selected
case, we replace the added spatial alternative with one irrelevant option
(\emph{quartz}, \emph{violin}, or \emph{lantern}) in the same ternary template
and rerun the exhaustive permutation protocol on matched samples. Because
sample-wise robust accuracy requires every option order to be correct, the
spatial and irrelevant conditions face the same permutation burden, reducing
position-bias explanations. This control isolates the key question: whether
spatial word meaning, not merely an added option position, drives the failure.

\begin{figure}[t]
\centering
\includegraphics[width=\columnwidth]{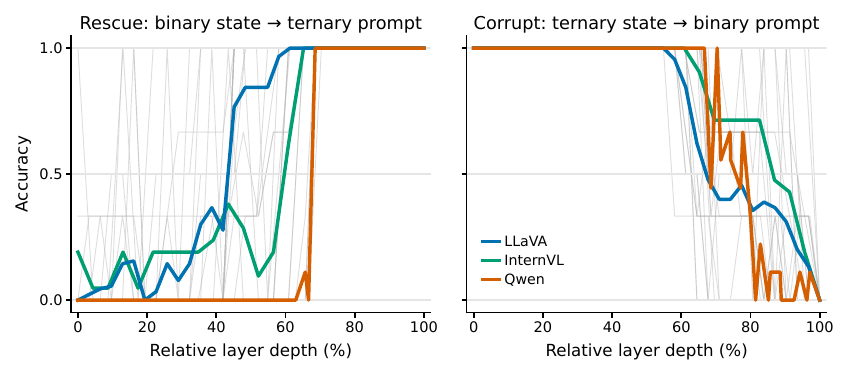}
\caption{\textbf{Last-token state exchange on the strict matched subset.}
Left: copying the binary last-token residual-stream vector into the ternary prompt restores
the answer in later layers. Right: copying the ternary last-token residual-stream vector into
the binary prompt corrupts the otherwise correct answer. Thin lines represent
selected BSTF cases; thick lines denote family means. The dashed gray line marks the
unpatched target baseline in each direction.}
\label{fig:patching_summary}
\end{figure}

Figure~\ref{fig:irrelevant_control} answers the specificity test: the drop from
irrelevant options is far smaller than the drop from spatial options. The mean
drop is 16.7 percentage points for irrelevant words, versus 97.8 percentage
points for spatial alternatives on the same selected BSTF cases. The error
distribution points to the same source: 97.7\% of spatial-option errors select
the newly added spatial alternative, whereas only 1.3\% of irrelevant-option
errors select the irrelevant word. Per-word and per-model controls are in
Appendix~\ref{app:irrelevant_details}. Thus step one confirms the specificity
claim: spatial relation words, not irrelevant added words, create the dominant
bias.

\paragraph{Ternary-to-four-way extension.}
Still in the first step, we test whether the spatial-word bias is limited to
the ternary diagnostic or persists in the full four-label setting. For each
selected BSTF case, the four-way extension adds the remaining spatial relation
and evaluates it under the same sample-wise robust scoring. This extension gives
the same conclusion: the binary-to-ternary cases are a clean diagnostic tool,
not the boundary of the phenomenon, and the same spatial-option bias also
appears in four-way prompts. Later, Section~\ref{sec:repair} tests whether
LLM-only DPO can repair this four-way setting as well. Thus step one establishes
the bias and its scope: spatial answer words create a structured option-set
effect. We next localize where this LLM-side bias can switch the answer.

\begin{figure*}[t!]
\centering
\includegraphics[width=\textwidth]{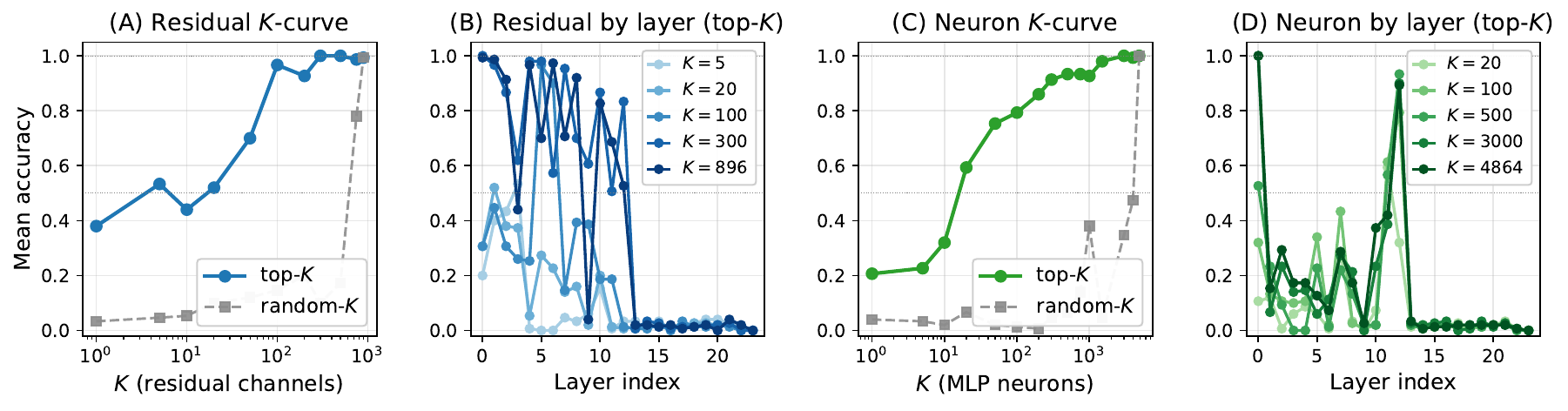}
\caption{\textbf{Top-$K$ component knockout at the added-option position on
InternVL2-1B.}
\textbf{(A,~C)}~Best-layer accuracy over $K$ for residual-stream channels and MLP
neurons, compared against size-matched random controls.
\textbf{(B,~D)}~Layer profiles for representative $K$ values. The repair effect
is strong in the early-to-mid stack but vanishes after layer~12.}
\label{fig:component_example}
\end{figure*}

\subsection{Layer-Wise Activation Patching Localizes the Decision Transition}
\label{sec:patching}

The second step is layer-level localization: at which layers can an LLM state
switch the selected answer? We exchange the entire last-token residual-stream
vector between strictly matched binary-correct and ternary-wrong prompts, layer
by layer. Filtering details are in Appendix~\ref{app:component_control}. This
isolates a layer-level answer-switching test: if a copied state switches the
answer, then that layer's state carries decision-relevant signal.

The layer-level diagnostic gives the second conclusion: whole-state
interventions become decisive in the mid-to-late LLM stack.
Figure~\ref{fig:patching_summary}
shows a sharply symmetric transition. Injecting the binary last-token vector
rescues the ternary failure, while injecting the ternary last-token vector
corrupts an otherwise correct binary answer. Across models, rescue reaches the
all-correct plateau at about 60\% relative network depth on average, while
corruption first moves clean accuracy below 1.00 at about 64\%. This diagnostic
does not map a complete circuit, but it localizes a causal readout transition:
mid-to-late last-token residual-stream states are sufficient to switch the
selected answer in both directions.

\subsection{Sparse Diagnostics of Added-Option Residual Channels and MLP Neurons}
\label{sec:component_control}

\begin{figure}[t]
\centering
\includegraphics[width=.94\columnwidth]{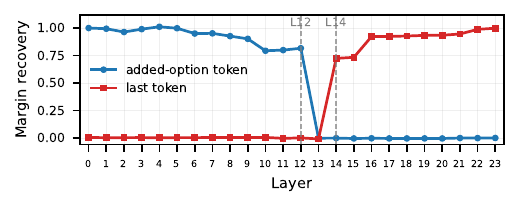}
\caption{\textbf{Layer-wise patching recovery at two token positions
(InternVL2-1B).} Patching the clean state into the failing prompt restores
the decision margin at the added-option token up to L12, and at the last
token from L14 onward.}
\label{fig:process_handoff}
\end{figure}

The third step is local component localization: can a small set of
added-option components modulate the error? We compare spatial-added ternary
prompts with matched irrelevant-option prompts, rank residual-stream channels
and MLP neurons by their activation differences at the added-option token, and
zero the top-\(K\) components. This tests whether the bias has local
added-option carriers, rather than only a diffuse whole-state signature
(Appendix~\ref{app:component_control}).

The sparse diagnostic gives the third conclusion: the error can be modulated by
a sparse added-option signal. Figure~\ref{fig:component_example} shows
InternVL2-1B as a worked example. Zeroing only 100 residual-stream channels at
layer~5 (\(11.2\%\) of the residual-stream width) or 200 MLP neurons at
layer~12 (\(4.1\%\) of the MLP width) recovers above 0.80 mean accuracy from
the zero baseline. The effect is also depth-limited: after layer~13
(\(57\%\) relative depth), even full residual-channel knockout at the added-option position no longer
restores accuracy (Appendix~\ref{app:component_layer_profile}). Thus step three identifies
local leverage points: added-option residual channels and MLP neurons can
strongly modulate the selected error before the signal leaves that token
position.

\textbf{From the added-option token to the decision position.} Where
does the added-option signal go after layer~13? Patching whole states at
both positions on a held-out InternVL2-1B case
(Figure~\ref{fig:process_handoff}) shows a handoff. The added-option token
restores the decision margin only up to L12, the last token only from L14
onward, and resetting the L14 last-token state removes 63.8\% of an L12
rescue. This is a coarse-grained, single-model account rather than a
circuit; the protocol and controls are given in
Appendix~\ref{app:process_level}.

The selected components also hold up beyond the discovery data.
Frozen component sets replayed on a held-out case recover far more of the
decision margin than equal-size random sets, most high-ranked components are
selective for which relation word was added, and relaxing the BSTF selection
gate leaves these findings unchanged
(Appendices~\ref{app:component_validation_full}
and~\ref{app:threshold_sensitivity}).

Together, these diagnostics locate spatial lexical bias at three resolutions.
First, the irrelevant-option and four-way tests confirm what the bias is:
spatial answer words create a structured option-set effect. Second, activation
patching locates where the decision becomes switchable: mid-to-late
last-token residual-stream states. Third, sparse knockout identifies local
leverage points: added-option residual channels and MLP neurons before the
signal leaves that token position.
The probe results of Section~\ref{sec:spatial_probe} provide
\emph{representational} evidence that the relation stays available, while
the patching and knockout experiments add \emph{causal control} at
specific token positions, depths, and components.
We do not claim a complete circuit or a
unique mechanistic flow, but these results do localize concrete LLM-side sites
where the spatial lexical bias can be detected and intervened on. This motivates
the LLM-only repair test in Section~\ref{sec:repair}.

\section{LLM-Only DPO Mitigation}
\label{sec:repair}

\providecommand{\dpoaccnum}[1]{\makebox[2.65em][r]{#1}}
\providecommand{\dpodeltanum}[1]{\makebox[3.05em][r]{#1}}
\providecommand{\dpodeltabox}[2]{\begingroup\setlength{\fboxsep}{0.4pt}\colorbox{#1}{\dpodeltanum{#2}}\endgroup}
\providecommand{\dpostackcell}[2]{\begin{tabular}[c]{@{}c@{}}#1\\[-0.15ex]#2\end{tabular}}
\providecommand{\dpomodelcell}[1]{\begin{tabular}[c]{@{}l@{}}#1\end{tabular}}

\begin{table*}[t!]
\centering
\scriptsize
\setlength{\tabcolsep}{2.0pt}
\resizebox{\textwidth}{!}{%
\begin{tabular}{ll|ccc|ccc|c}
\toprule
\textbf{Model} & \textbf{Row} & \multicolumn{3}{c|}{\textbf{Controlled Synthetic}} & \multicolumn{3}{c|}{\textbf{Public}} & \textbf{Model Avg.} \\
& & \textbf{CS} & \textbf{DR} & \textbf{DO} & \textbf{WU} & \textbf{SM} & \textbf{VS} & \textbf{Avg.} \\
\midrule
\rowcolor{gray!16}\multicolumn{9}{l}{\textit{LLaVA family}}\\
\dpomodelcell{LLaVA-1.5-7B} & \dpostackcell{DPO}{$\Delta$} & \dpostackcell{\dpoaccnum{93.8}/\dpoaccnum{90.3}/\dpoaccnum{80.4}}{\dpodeltabox{green!32}{+72.5}/\dpodeltabox{green!32}{+90.0}/\dpodeltabox{green!32}{+80.4}} & \dpostackcell{\dpoaccnum{90.3}/\dpoaccnum{83.9}/\dpoaccnum{76.5}}{\dpodeltabox{green!32}{+54.6}/\dpodeltabox{green!32}{+82.4}/\dpodeltabox{green!32}{+76.5}} & \dpostackcell{\dpoaccnum{88.8}/\dpoaccnum{81.2}/\dpoaccnum{68.9}}{\dpodeltabox{green!32}{+34.2}/\dpodeltabox{green!32}{+58.2}/\dpodeltabox{green!32}{+66.1}} & \dpostackcell{\dpoaccnum{78.5}/\dpoaccnum{67.2}/\dpoaccnum{56.0}}{\dpodeltabox{green!32}{+38.9}/\dpodeltabox{green!32}{+50.6}/\dpodeltabox{green!32}{+49.6}} & \dpostackcell{\dpoaccnum{49.6}/\dpoaccnum{33.6}/\dpoaccnum{18.8}}{\dpodeltabox{green!32}{+33.4}/\dpodeltabox{green!32}{+28.2}/\dpodeltabox{green!24}{+16.2}} & \dpostackcell{\dpoaccnum{49.0}/\dpoaccnum{28.4}/\dpoaccnum{16.6}}{\dpodeltabox{green!16}{+12.5}/\dpodeltabox{green!16}{+11.6}/\dpodeltabox{green!16}{+7.7}} & \dpostackcell{\dpoaccnum{75.0}/\dpoaccnum{64.1}/\dpoaccnum{52.9}}{\dpodeltabox{green!32}{+41.0}/\dpodeltabox{green!32}{+53.5}/\dpodeltabox{green!32}{+49.4}} \\
\addlinespace[2pt]
\dpomodelcell{LLaVA-v1.6-Vicuna-7B} & \dpostackcell{DPO}{$\Delta$} & \dpostackcell{\dpoaccnum{100.0}/\dpoaccnum{98.6}/\dpoaccnum{69.4}}{\dpodeltabox{green!32}{+27.7}/\dpodeltabox{green!32}{+40.5}/\dpodeltabox{green!24}{+19.4}} & \dpostackcell{\dpoaccnum{99.5}/\dpoaccnum{97.8}/\dpoaccnum{92.2}}{\dpodeltabox{green!24}{+16.4}/\dpodeltabox{green!32}{+32.4}/\dpodeltabox{green!32}{+47.4}} & \dpostackcell{\dpoaccnum{100.0}/\dpoaccnum{99.8}/\dpoaccnum{98.9}}{\dpodeltabox{green!24}{+16.2}/\dpodeltabox{green!32}{+41.8}/\dpodeltabox{green!32}{+71.0}} & \dpostackcell{\dpoaccnum{95.0}/\dpoaccnum{82.1}/\dpoaccnum{59.6}}{\dpodeltabox{green!24}{+20.0}/\dpodeltabox{green!32}{+44.7}/\dpodeltabox{green!32}{+33.3}} & \dpostackcell{\dpoaccnum{70.7}/\dpoaccnum{51.6}/\dpoaccnum{32.8}}{\dpodeltabox{green!16}{+13.8}/\dpodeltabox{green!32}{+25.2}/\dpodeltabox{green!24}{+23.1}} & \dpostackcell{\dpoaccnum{65.9}/\dpoaccnum{42.1}/\dpoaccnum{23.9}}{\dpodeltabox{green!16}{+10.1}/\dpodeltabox{green!16}{+10.9}/\dpodeltabox{green!9}{+6.0}} & \dpostackcell{\dpoaccnum{88.5}/\dpoaccnum{78.7}/\dpoaccnum{62.8}}{\dpodeltabox{green!24}{+17.4}/\dpodeltabox{green!32}{+32.6}/\dpodeltabox{green!32}{+33.4}} \\
\addlinespace[2pt]
\dpomodelcell{LLaVA-v1.6-Mistral-7B} & \dpostackcell{DPO}{$\Delta$} & \dpostackcell{\dpoaccnum{99.8}/\dpoaccnum{90.9}/\dpoaccnum{90.3}}{\dpodeltabox{green!32}{+52.4}/\dpodeltabox{green!32}{+74.2}/\dpodeltabox{green!32}{+90.3}} & \dpostackcell{\dpoaccnum{98.5}/\dpoaccnum{100.0}/\dpoaccnum{100.0}}{\dpodeltabox{green!32}{+56.5}/\dpodeltabox{green!32}{+80.9}/\dpodeltabox{green!32}{+100.0}} & \dpostackcell{\dpoaccnum{97.7}/\dpoaccnum{100.0}/\dpoaccnum{100.0}}{\dpodeltabox{green!32}{+43.4}/\dpodeltabox{green!32}{+67.3}/\dpodeltabox{green!32}{+97.4}} & \dpostackcell{\dpoaccnum{98.0}/\dpoaccnum{93.6}/\dpoaccnum{84.2}}{\dpodeltabox{green!32}{+36.2}/\dpodeltabox{green!32}{+57.0}/\dpodeltabox{green!32}{+68.0}} & \dpostackcell{\dpoaccnum{70.1}/\dpoaccnum{58.7}/\dpoaccnum{48.2}}{\dpodeltabox{green!16}{+8.7}/\dpodeltabox{green!24}{+21.4}/\dpodeltabox{green!32}{+32.6}} & \dpostackcell{\dpoaccnum{58.6}/\dpoaccnum{43.1}/\dpoaccnum{29.5}}{\dpodeltabox{green!16}{+14.9}/\dpodeltabox{green!24}{+22.3}/\dpodeltabox{green!24}{+20.1}} & \dpostackcell{\dpoaccnum{87.1}/\dpoaccnum{81.0}/\dpoaccnum{75.3}}{\dpodeltabox{green!32}{+35.3}/\dpodeltabox{green!32}{+53.9}/\dpodeltabox{green!32}{+68.1}} \\
\addlinespace[2pt]
\addlinespace
\rowcolor{gray!16}\multicolumn{9}{l}{\textit{Qwen family}}\\
\dpomodelcell{Qwen2-VL-2B} & \dpostackcell{DPO}{$\Delta$} & \dpostackcell{\dpoaccnum{100.0}/\dpoaccnum{100.0}/\dpoaccnum{100.0}}{\dpodeltabox{green!32}{+25.8}/\dpodeltabox{green!32}{+38.8}/\dpodeltabox{green!32}{+48.5}} & \dpostackcell{\dpoaccnum{98.8}/\dpoaccnum{90.1}/\dpoaccnum{64.3}}{\dpodeltabox{green!16}{+14.6}/\dpodeltabox{green!32}{+33.0}/\dpodeltabox{green!32}{+46.5}} & \dpostackcell{\dpoaccnum{88.1}/\dpoaccnum{69.6}/\dpoaccnum{35.6}}{\dpodeltabox{green!24}{+21.7}/\dpodeltabox{green!32}{+40.3}/\dpodeltabox{green!32}{+25.2}} & \dpostackcell{\dpoaccnum{97.6}/\dpoaccnum{92.9}/\dpoaccnum{84.5}}{\dpodeltabox{green!24}{+19.5}/\dpodeltabox{green!32}{+29.0}/\dpodeltabox{green!32}{+36.7}} & \dpostackcell{\dpoaccnum{72.4}/\dpoaccnum{57.5}/\dpoaccnum{41.9}}{\dpodeltabox{gray!10}{-1.3}/\dpodeltabox{green!9}{+2.9}/\dpodeltabox{green!9}{+2.1}} & \dpostackcell{\dpoaccnum{67.9}/\dpoaccnum{48.9}/\dpoaccnum{33.6}}{\dpodeltabox{green!9}{+6.2}/\dpodeltabox{green!16}{+9.2}/\dpodeltabox{green!16}{+10.1}} & \dpostackcell{\dpoaccnum{87.5}/\dpoaccnum{76.5}/\dpoaccnum{60.0}}{\dpodeltabox{green!16}{+14.4}/\dpodeltabox{green!32}{+25.5}/\dpodeltabox{green!32}{+28.2}} \\
\addlinespace[2pt]
\dpomodelcell{Qwen2-VL-7B} & \dpostackcell{DPO}{$\Delta$} & \dpostackcell{\dpoaccnum{100.0}/\dpoaccnum{100.0}/\dpoaccnum{100.0}}{\dpodeltabox{gray!10}{+0.0}/\dpodeltabox{gray!10}{+0.0}/\dpodeltabox{gray!10}{-0.1}} & \dpostackcell{\dpoaccnum{100.0}/\dpoaccnum{99.9}/\dpoaccnum{100.0}}{\dpodeltabox{gray!10}{+0.1}/\dpodeltabox{gray!10}{+1.5}/\dpodeltabox{green!9}{+2.2}} & \dpostackcell{\dpoaccnum{99.1}/\dpoaccnum{91.9}/\dpoaccnum{87.8}}{\dpodeltabox{green!9}{+3.1}/\dpodeltabox{green!16}{+8.3}/\dpodeltabox{green!16}{+9.3}} & \dpostackcell{\dpoaccnum{99.9}/\dpoaccnum{99.4}/\dpoaccnum{98.0}}{\dpodeltabox{gray!10}{+0.9}/\dpodeltabox{gray!10}{+1.9}/\dpodeltabox{green!9}{+3.9}} & \dpostackcell{\dpoaccnum{81.9}/\dpoaccnum{67.3}/\dpoaccnum{55.7}}{\dpodeltabox{gray!10}{-0.7}/\dpodeltabox{gray!10}{-0.3}/\dpodeltabox{gray!10}{+0.5}} & \dpostackcell{\dpoaccnum{69.7}/\dpoaccnum{49.0}/\dpoaccnum{35.8}}{\dpodeltabox{gray!10}{-1.1}/\dpodeltabox{gray!10}{+0.0}/\dpodeltabox{green!9}{+3.0}} & \dpostackcell{\dpoaccnum{91.8}/\dpoaccnum{84.6}/\dpoaccnum{79.5}}{\dpodeltabox{gray!10}{+0.4}/\dpodeltabox{gray!10}{+1.9}/\dpodeltabox{green!9}{+3.1}} \\
\addlinespace[2pt]
\dpomodelcell{Qwen3-VL-2B} & \dpostackcell{DPO}{$\Delta$} & \dpostackcell{\dpoaccnum{100.0}/\dpoaccnum{100.0}/\dpoaccnum{100.0}}{\dpodeltabox{green!24}{+16.1}/\dpodeltabox{green!32}{+26.7}/\dpodeltabox{green!32}{+35.6}} & \dpostackcell{\dpoaccnum{99.4}/\dpoaccnum{94.6}/\dpoaccnum{88.9}}{\dpodeltabox{green!16}{+13.0}/\dpodeltabox{green!24}{+18.1}/\dpodeltabox{green!16}{+14.0}} & \dpostackcell{\dpoaccnum{99.0}/\dpoaccnum{94.6}/\dpoaccnum{89.6}}{\dpodeltabox{green!16}{+14.7}/\dpodeltabox{green!24}{+17.8}/\dpodeltabox{green!16}{+14.6}} & \dpostackcell{\dpoaccnum{99.2}/\dpoaccnum{95.9}/\dpoaccnum{90.7}}{\dpodeltabox{green!9}{+5.6}/\dpodeltabox{green!16}{+12.7}/\dpodeltabox{green!24}{+19.1}} & \dpostackcell{\dpoaccnum{87.2}/\dpoaccnum{76.6}/\dpoaccnum{66.0}}{\dpodeltabox{gray!10}{+1.2}/\dpodeltabox{green!9}{+2.0}/\dpodeltabox{green!9}{+4.1}} & \dpostackcell{\dpoaccnum{75.3}/\dpoaccnum{58.4}/\dpoaccnum{43.9}}{\dpodeltabox{green!9}{+5.9}/\dpodeltabox{green!16}{+7.5}/\dpodeltabox{green!16}{+9.3}} & \dpostackcell{\dpoaccnum{93.3}/\dpoaccnum{86.7}/\dpoaccnum{79.9}}{\dpodeltabox{green!16}{+9.4}/\dpodeltabox{green!16}{+14.1}/\dpodeltabox{green!24}{+16.1}} \\
\addlinespace[2pt]
\dpomodelcell{Qwen3-VL-8B} & \dpostackcell{DPO}{$\Delta$} & \dpostackcell{\dpoaccnum{99.8}/\dpoaccnum{99.3}/\dpoaccnum{96.5}}{\dpodeltabox{green!9}{+7.0}/\dpodeltabox{green!24}{+15.5}/\dpodeltabox{green!24}{+21.5}} & \dpostackcell{\dpoaccnum{97.3}/\dpoaccnum{93.2}/\dpoaccnum{87.9}}{\dpodeltabox{green!9}{+5.2}/\dpodeltabox{green!16}{+9.4}/\dpodeltabox{green!16}{+12.4}} & \dpostackcell{\dpoaccnum{98.6}/\dpoaccnum{96.2}/\dpoaccnum{94.3}}{\dpodeltabox{green!9}{+3.9}/\dpodeltabox{green!16}{+10.3}/\dpodeltabox{green!24}{+23.6}} & \dpostackcell{\dpoaccnum{98.6}/\dpoaccnum{96.9}/\dpoaccnum{88.7}}{\dpodeltabox{green!9}{+2.2}/\dpodeltabox{green!9}{+4.7}/\dpodeltabox{green!9}{+4.9}} & \dpostackcell{\dpoaccnum{91.8}/\dpoaccnum{83.9}/\dpoaccnum{76.0}}{\dpodeltabox{gray!10}{+1.3}/\dpodeltabox{green!9}{+2.2}/\dpodeltabox{green!9}{+4.6}} & \dpostackcell{\dpoaccnum{75.8}/\dpoaccnum{57.7}/\dpoaccnum{42.6}}{\dpodeltabox{green!9}{+2.4}/\dpodeltabox{green!9}{+3.2}/\dpodeltabox{green!9}{+4.5}} & \dpostackcell{\dpoaccnum{93.6}/\dpoaccnum{87.9}/\dpoaccnum{81.0}}{\dpodeltabox{green!9}{+3.7}/\dpodeltabox{green!16}{+7.5}/\dpodeltabox{green!16}{+11.9}} \\
\addlinespace[2pt]
\addlinespace
\rowcolor{gray!16}\multicolumn{9}{l}{\textit{InternVL family}}\\
\dpomodelcell{InternVL2-1B} & \dpostackcell{DPO}{$\Delta$} & \dpostackcell{\dpoaccnum{99.6}/\dpoaccnum{92.5}/\dpoaccnum{67.8}}{\dpodeltabox{green!32}{+30.4}/\dpodeltabox{green!32}{+57.9}/\dpodeltabox{green!32}{+42.8}} & \dpostackcell{\dpoaccnum{91.5}/\dpoaccnum{74.4}/\dpoaccnum{52.3}}{\dpodeltabox{green!24}{+24.6}/\dpodeltabox{green!32}{+40.3}/\dpodeltabox{green!32}{+27.5}} & \dpostackcell{\dpoaccnum{89.6}/\dpoaccnum{65.7}/\dpoaccnum{49.6}}{\dpodeltabox{green!32}{+33.2}/\dpodeltabox{green!32}{+30.7}/\dpodeltabox{green!24}{+24.6}} & \dpostackcell{\dpoaccnum{87.4}/\dpoaccnum{67.5}/\dpoaccnum{32.8}}{\dpodeltabox{green!24}{+20.6}/\dpodeltabox{green!32}{+30.2}/\dpodeltabox{green!24}{+15.6}} & \dpostackcell{\dpoaccnum{59.4}/\dpoaccnum{38.8}/\dpoaccnum{24.9}}{\dpodeltabox{red!10}{-3.0}/\dpodeltabox{green!9}{+2.6}/\dpodeltabox{green!9}{+4.7}} & \dpostackcell{\dpoaccnum{59.2}/\dpoaccnum{34.1}/\dpoaccnum{15.0}}{\dpodeltabox{green!9}{+3.1}/\dpodeltabox{green!16}{+8.8}/\dpodeltabox{green!9}{+6.1}} & \dpostackcell{\dpoaccnum{81.1}/\dpoaccnum{62.2}/\dpoaccnum{40.4}}{\dpodeltabox{green!24}{+18.2}/\dpodeltabox{green!32}{+28.4}/\dpodeltabox{green!24}{+20.2}} \\
\addlinespace[2pt]
\dpomodelcell{InternVL2.5-1B} & \dpostackcell{DPO}{$\Delta$} & \dpostackcell{\dpoaccnum{97.2}/\dpoaccnum{90.0}/\dpoaccnum{75.1}}{\dpodeltabox{green!24}{+24.9}/\dpodeltabox{green!32}{+32.8}/\dpodeltabox{green!32}{+31.6}} & \dpostackcell{\dpoaccnum{98.8}/\dpoaccnum{96.0}/\dpoaccnum{91.4}}{\dpodeltabox{green!16}{+12.8}/\dpodeltabox{green!32}{+41.0}/\dpodeltabox{green!32}{+42.8}} & \dpostackcell{\dpoaccnum{98.9}/\dpoaccnum{90.4}/\dpoaccnum{82.5}}{\dpodeltabox{green!9}{+2.8}/\dpodeltabox{green!32}{+34.7}/\dpodeltabox{green!32}{+50.4}} & \dpostackcell{\dpoaccnum{88.8}/\dpoaccnum{77.0}/\dpoaccnum{73.4}}{\dpodeltabox{green!24}{+17.8}/\dpodeltabox{green!32}{+29.8}/\dpodeltabox{green!32}{+34.2}} & \dpostackcell{\dpoaccnum{54.9}/\dpoaccnum{35.4}/\dpoaccnum{28.6}}{\dpodeltabox{gray!10}{+0.1}/\dpodeltabox{green!16}{+8.8}/\dpodeltabox{green!9}{+3.6}} & \dpostackcell{\dpoaccnum{50.8}/\dpoaccnum{31.9}/\dpoaccnum{22.0}}{\dpodeltabox{green!9}{+3.0}/\dpodeltabox{green!16}{+10.0}/\dpodeltabox{green!16}{+8.3}} & \dpostackcell{\dpoaccnum{81.6}/\dpoaccnum{70.1}/\dpoaccnum{62.2}}{\dpodeltabox{green!16}{+10.2}/\dpodeltabox{green!32}{+26.2}/\dpodeltabox{green!32}{+28.5}} \\
\addlinespace[2pt]
\bottomrule
\end{tabular}%
}
\caption{\textbf{Strict sample-wise DPO accuracy and change by model and dataset.} Each model row stacks post-DPO binary/ternary/four-way robust accuracy (\%) above the post-DPO minus base change (percentage points). Color is applied only to the change line and separately for binary/ternary/four-way values.}
\label{tab:dpo_samplewise_model_dataset_repair}
\end{table*}

Our DPO experiment asks whether the LLM-side bias can be repaired with almost no
new visual supervision. The setup is intentionally minimal: \textbf{we build
preference data from one single torus--pyramid object pair}, and this
\textbf{introduces almost no new visual information}. We then train only
LoRA-DPO adapters~\citep{hu2022lora} in the LLM decoder, with the vision encoder and projector kept
frozen. Thus this section tests a simple claim: if this minimal LLM-only update
repairs the failure, spatial lexical bias is an actionable language-side
decision bias.

\begin{figure}[h!]
\centering
\setlength{\tabcolsep}{1pt}
\begin{tabular}{cccc}
\includegraphics[width=.235\columnwidth]{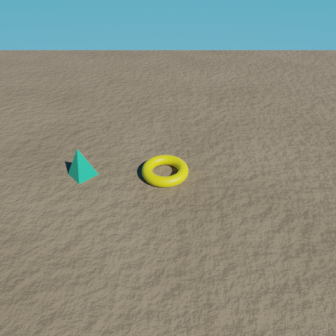} &
\includegraphics[width=.235\columnwidth]{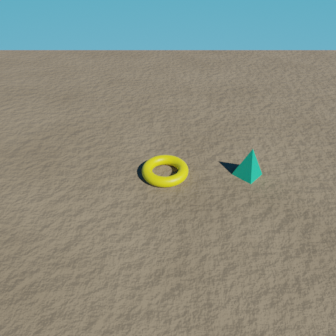} &
\includegraphics[width=.235\columnwidth]{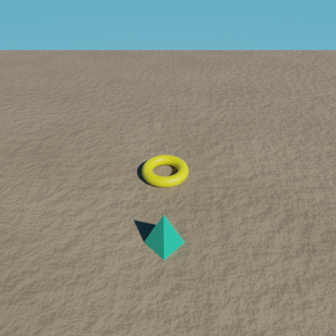} &
\includegraphics[width=.235\columnwidth]{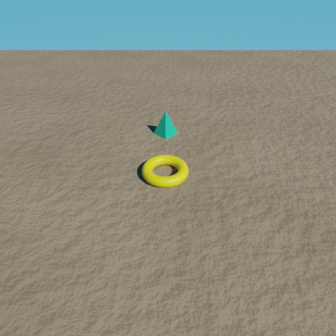} \\
{\scriptsize left} & {\scriptsize right} & {\scriptsize front} & {\scriptsize behind}
\end{tabular}
\caption{\textbf{DPO training source.}
Four relation variants from the \textbf{single torus--pyramid object pair} used
for preference training.}
\label{fig:dpo_training_source}
\end{figure}

For each of the nine models, we train three LoRA-DPO adapters on the tiny
disjoint Pyramid--Torus--Outdoor source
(Figure~\ref{fig:dpo_training_source};
Appendix~\ref{app:dpo_training_source}). All base parameters are frozen, and
LoRA is inserted only into the LLM decoder; training details are in
Appendix~\ref{app:dpo_training_setup}. The source is deliberately too small to
teach broad new visual knowledge, so success tests whether LLM-side option
dynamics can be corrected.

The preference data are also minimal: they supervise answer letters, not
free-form rationales. Chosen and rejected letters are balanced within each scene
and option ordering, so no spatial word or answer letter receives a net
preference advantage. This keeps the repair focused on the image-conditioned
relation, not a lexical answer-word shortcut
(Appendix~\ref{app:dpo_training_setup}).

\begin{figure}[t]
\centering
\includegraphics[width=.9\columnwidth]{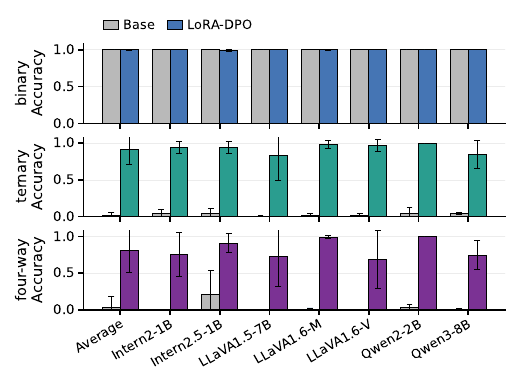}
\caption{\textbf{DPO repair on the 55 selected BSTF cases.}
Sample-wise robust accuracy before and after LoRA-DPO; bars show model averages
and per-model means, with case-level standard deviations.}
\label{fig:dpo_repair}
\end{figure}

We first test whether this single-pair LLM-only repair works on the most extreme
BSTF failures. For the seven models represented in the 55-case set, we evaluate
each adapter on its model's subset of the selected BSTF cases, using the same
sample-wise robust metric as Section~\ref{sec:highhigh}. The repair effect is
large and selective across prompt sizes (Figure~\ref{fig:dpo_repair}): binary
accuracy is preserved ($1.000\!\to\!0.999$), ternary accuracy rises from $0.022$
to $0.917$ (\(\Delta_{\rm tern}=+0.894\)), and four-way accuracy rises from
$0.038$ to $0.816$. The ternary repair effect is positive for every individual
model (\(+0.802\) to \(+0.968\)). Thus, on the hardest diagnostic cases, a
single-pair LLM-only update repairs the spatial lexical-bias failure while
preserving the binary behavior.

We then test whether the same minimal repair generalizes beyond the selected
BSTF cases. We run the adapted models on the full six-dataset sweep: the three
controlled synthetic evaluation sets plus
WhatsUp-B~\citep{kamath-etal-2023-whats},
SpatialMQA-Direct~\citep{liu-etal-2025-multimodal-large}, and
VSR~\citep{liu-etal-2023-visual} (Appendix~\ref{app:dataset_details}).
Mean sample-wise robust accuracy rises from
$0.649/0.418/0.278$ to $0.849/0.744/0.621$ across binary/ternary/four-way
prompts. At the strongest four-way setting, gains reach 100.0 points on
controlled synthetic data and 68.0, 32.6, and 20.1 points on WhatsUp-B,
SpatialMQA-Direct, and VSR, respectively;
Table~\ref{tab:dpo_samplewise_model_dataset_repair} gives the
model--dataset breakdown. We interpret these results conservatively: this repair
is not by itself a unique proof of the mechanism. Still, because it uses only
one object pair and freezes the visual pathway, its transfer supports the main
claim: spatial lexical bias is important, LLM-side, and repairable.

\textbf{DPO is not the only effective repair.} SFT on the same pairs,
with the same frozen vision encoder and LoRA setup, achieves similar
improvements, matching DPO within 0.038 ternary accuracy on six of seven
models with binary preserved. Prompting does not repair the failure: three instruction-only variants
reach at most 10.7\%, with errors still selecting the added word
(Appendix~\ref{app:sft_prompt}).

\textbf{The repair transfers to relations absent from training.} On
What'sUp-A \emph{on}/\emph{under} images (neither relation trained), base
accuracy declines as options grow (0.787/0.770/0.742) while the repair
gain grows (+1.0/+5.0/+7.0 points), and no model declines in the
four-option setting despite the trained \emph{left}/\emph{right} appearing
as competitors (Appendix~\ref{app:unseen_relations}).

\textbf{General capabilities are preserved.} Across MME~\citep{NEURIPS2025_d79a27cf},
MMBench~\citep{10.1007/978-3-031-72658-3_13}, COCO
captioning~\citep{10.1007/978-3-319-10602-1_48,chen2015microsoftcococaptionsdata},
MMLU~\citep{hendrycks2021measuring}, and
IFEval~\citep{zhou2023instructionfollowingevaluationlargelanguage} on all
seven repaired models, changes
are small and two-sided (MME $-0.46$ points macro; IFEval $+1.40$ with no
model degrading; Appendix~\ref{app:capability_checks}).

\section{Conclusion}
\label{sec:conclusion}

This work identifies \textit{spatial lexical bias} as a language-side failure
mode in MLLM spatial multiple-choice reasoning. Across open-weight MLLMs,
BSTF cases show that a model can solve a binary spatial contrast yet collapse
when a plausible third spatial word is added. Attention stays on the same object
regions, and last-token probes recover the true relation nearly perfectly, even
from early layers; the collapse is therefore not primarily visual information
loss.

Our diagnostics locate the failure on the LLM side. Irrelevant-option and
four-way controls show structured spatial-word interactions, not a choice-count
effect. Activation patching places the answer switch in mid-to-late residual
states, and sparse knockout shows that added-option residual/MLP components can
restore selected failures. Thus, spatial lexical bias is mechanistically
localizable.

LLM-only LoRA-DPO makes the bias actionable. Training only the language backbone
on a single-pair synthetic source nearly resolves the selected BSTF set while
preserving binary accuracy (\(0.022/0.038{\to}0.917/0.816\) for
ternary/four-way) and transfers beyond the mined cases. Spatial lexical bias is therefore a concrete
LLM-side repair target.

\section*{Limitations}

This work isolates \textit{spatial lexical bias} in a deliberately controlled
spatial-relation setting. First, our DPO mitigation uses a very small synthetic
source built from a single object pair. This design is useful because it tests
whether the bias can be repaired with almost no new visual information, but it
does not cover the full visual, linguistic, and compositional diversity of
open-world spatial reasoning.

Second, our diagnostic setting focuses on four-direction multiple-choice
relations. This control helps separate lexical preference from visual grounding
failure, but errors on richer in-the-wild benchmarks may involve additional
mechanisms, including object recognition, scene complexity, relation ambiguity,
and instruction-following failures. Our interventions also localize manipulable
LLM-side states and components, but they do not establish a complete circuit.
Extending the analysis and repair method to richer scenes, broader spatial
relations, and open-ended outputs remains an important direction for future
work.
Our option-free one-word experiments show that the bias does not
vanish when the displayed options are removed, but the preferences that
surface in that setting differ from the option-set effect studied here
(Appendix~\ref{app:open_ended}). The mechanistic account of this paper does
not transfer directly to open-ended generation; we leave this to future
work.

This work establishes lexical bias only for spatial reasoning. Whether
similar option-induced biases arise in other tasks, such as counting,
color, or questions with ambiguous answers, requires further experimental
validation.

\section*{Acknowledgments}
This work was supported in part by MEXT KAKENHI Grant-in-Aid for Transformative Research Areas (A), “Foundation of Machine Learning Physics” (Grant Number JP22H05117) (TO), and by JSPS
KAKENHI Grant Number JP26K25543 (QL). We used ABCI 3.0 provided by AIST and AIST Solutions with support from
``ABCI 3.0 Development Acceleration Use''. In this research work, we
used the ``mdx: a platform for building data-empowered society''.

We used AI assistants for translation, polishing, and editing of the
writing, and for implementing and debugging the experimental and analysis
code. All research ideas, experimental designs, and claims were developed
and verified by the authors.

\bibliography{references}

\clearpage
\appendix
\setcounter{figure}{0}
\setcounter{table}{0}
\renewcommand{\thefigure}{A\arabic{figure}}
\renewcommand{\thetable}{A\arabic{table}}
\renewcommand{\theHfigure}{appendix.\thefigure}
\renewcommand{\theHtable}{appendix.\thetable}

\section{Dataset Details}
\label{app:dataset_details}

Table~\ref{tab:datasets_overview} lists the six evaluation datasets and their
label counts after mapping to the four-relation label space.

\begin{table*}[t]
\centering
\small
\setlength{\tabcolsep}{5pt}
\begin{tabular*}{\textwidth}{@{\extracolsep{\fill}}lllllrrrr@{}}
\toprule
 & & & & & \multicolumn{4}{c}{\textbf{Label counts}} \\
\cmidrule(l){6-9}
\textbf{Dataset} & \textbf{Abbr.} & \textbf{Source} & \textbf{Imagery} & \textbf{Images} & \textbf{Left} & \textbf{Right} & \textbf{Front} & \textbf{Behind} \\
\midrule
WhatsUp-B~\citep{kamath-etal-2023-whats}         & WU & Public    & Synthetic & 408   & 102 & 102 & 102 & 102 \\
VSR~\citep{liu-etal-2023-visual}               & VS & Public    & Natural   & 1,197 & 318 & 298 & 384 & 357 \\
SpatialMQA-Direct~\citep{liu-etal-2025-multimodal-large} & SM & Public    & Natural   & 782   & 302 & 314 &  42 & 124 \\
\addlinespace
Sphere-Cube Raw    & CS & New       & Synthetic & 2,000 & 500 & 500 & 500 & 500 \\
Sphere-Dog Raw     & DR & New       & Synthetic & 2,000 & 500 & 500 & 500 & 500 \\
Sphere-Dog Outdoor & DO & New       & Synthetic & 2,000 & 500 & 500 & 500 & 500 \\
\bottomrule
\end{tabular*}
\caption{\textbf{Datasets used in this study.} The top three rows are public
benchmarks; the bottom three are newly rendered synthetic datasets with
balanced labels. For VSR, a single image can carry multiple spatial-relation
annotations for different object pairs, so the label counts (1{,}357) exceed
the number of unique images (1{,}197).}
\label{tab:datasets_overview}
\end{table*}

\paragraph{Public-benchmark subset construction.}
The three public-benchmark evaluation sets are subsets that we extract from
the original releases and project onto the four-relation label space. We document the
exact source identifier and selection rule used for each, so that the
counts in Table~\ref{tab:datasets_overview} are reproducible from the
upstream data.

\textbf{WhatsUp-B} uses the public Controlled Images B subset
(also called the controlled CLEVR subset) from
\citet{kamath-etal-2023-whats}\footnote{\url{https://github.com/amitakamath/whatsup_vlms}};
all 408 images (102 per relation) are kept.

\textbf{VSR} is built from the random split of
\citet{liu-etal-2023-visual}\footnote{\url{https://huggingface.co/datasets/cambridgeltl/vsr_random}}
(not the zero-shot
split). We keep only positively labeled captions
whose spatial relation maps to one of \emph{left}, \emph{right},
\emph{front}, or \emph{behind}, which yields 1{,}357 caption-level
samples over 1{,}197 unique images (8{,}302 samples are dropped for
non-four-way relations and 1{,}313 for being false captions).

\textbf{SpatialMQA-Direct} is built from
\citet{liu-etal-2025-multimodal-large}\footnote{\url{https://huggingface.co/datasets/liuziyan/SpatialMQA}}
by retaining only direct object-centric items whose question begins with
``Where is'' or ``Where are'' and whose true label is one of
\emph{left}, \emph{right}, \emph{front}, or \emph{behind}. From the
5{,}392 source samples this removes 3{,}634 non-direct questions, 949
non-four-way answers, and 27 duplicate images, leaving 782 samples.
The ``Direct'' qualifier denotes this surface-form filter.

\paragraph{Public artifact licenses and terms.}
We use the three public benchmarks only as evaluation sources after the
filtering above, and we do not redistribute public images. The official VSR
release describes the random split as image--caption pairs with true/false
spatial-relation labels; its Hugging Face dataset card lists CC-BY-4.0 for
the dataset records, while the project repository is Apache-2.0. The official
SpatialMQA release describes the dataset as a manually annotated
multiple-choice visual-question-answering benchmark over COCO2017
images~\citep{10.1007/978-3-319-10602-1_48}; its
Hugging Face dataset card lists CC-BY-4.0, while the project repository is
Apache-2.0. Both VSR and SpatialMQA use COCO images, whose official terms
state that COCO annotations and website content are CC-BY-4.0, while image
use follows the Flickr terms of use.
The official WhatsUp repository distributes code and datasets under an MIT
license and does not list a separate image-specific license in its README or
license file. We cite the creators of all three benchmark sources and retain
their original source identifiers in our preprocessing manifests.

\section{Controlled Synthetic Dataset Details}
\label{app:synthetic_datasets}

All three synthetic datasets are rendered with Blender's Cycles
path-tracing engine. Every dataset contains 500 scenes. Each scene is assigned
a unique random seed that determines the reference object's position within the
allowed area; the target object is then placed at a fixed distance
($d{=}1.6$ units) along each of the four cardinal directions. This yields
four images per scene and 2{,}000 images per dataset. Because the reference
position varies across scenes while all other visual elements stay fixed, the
500 scenes provide diverse spatial layouts without introducing confounding
visual variation.\footnote{The rendering scripts, assets, and generated
datasets are available at
\url{https://github.com/chuang-ma-ku/spatial-lexical-bias}.}

\paragraph{Assets.}
Blender is distributed under the GNU GPL, which applies to the Blender
application rather than artwork created with it.
The sphere and cube are built-in Blender primitives with solid-color materials.
The dog model is a low-polygon Shiba Inu mesh by Pat Siefring from Poly
Pizza\footnote{\url{https://poly.pizza/m/1sr1MDt9db5}} (CC-BY 3.0).
For the outdoor environment (Sphere-Dog Outdoor), the HDRI background is from
Poly Haven\footnote{\url{https://polyhaven.com/a/delta_2}} (CC0) and the
ground material is from Poly
Haven\footnote{\url{https://polyhaven.com/a/rocky_terrain_02}} (CC0).
Poly Haven releases its HDRIs, textures, and 3D models under CC0.
The ground plane and sphere geometries are generated procedurally by the
rendering procedure.

\paragraph{Sphere-Cube Raw.}
A red sphere and a blue cube are placed on a uniform gray checkerboard plane.
The camera is fixed above the scene, looking down at the scene center.
Lighting consists of a sun lamp and three area lights providing fill and
ambient illumination.
Because both objects are symmetric geometric primitives, this is the most
controlled setting---object recognition and geometric asymmetry are ruled out
as confounds.

\paragraph{Sphere-Dog Raw.}
The blue cube is replaced by a Shiba Inu mesh, introducing asymmetric object
geometry while keeping the checkerboard backdrop and lighting identical to
Sphere-Cube Raw. This lets us check whether models rely on object symmetry as
a shortcut for spatial judgments.

\paragraph{Sphere-Dog Outdoor.}
The same sphere--dog pair is placed in an outdoor grass environment with
natural HDRI lighting and a distant horizon. Visual complexity increases
(texture, shadows, depth cues) while the spatial configurations stay the same,
so this set tests whether the spatial reasoning patterns observed in the
simpler scenes still hold in a more naturalistic setting.

\subsection{DPO Training Source}
\label{app:dpo_training_source}

The LoRA-DPO repair uses a separate paired source,
Pyramid--Torus--Outdoor (Figure~\ref{fig:dpo_training_source}), with a yellow
torus and a teal pyramid in an outdoor sand/sea scene. It follows the same four-direction paired construction as the
evaluation sets, but is used only for preference training, validation, and
testing, not for BSTF selection or repair evaluation.

\paragraph{Spatial placement protocol.}
For each scene, the reference object is placed at a random position within the
allowed area (determined by the scene seed, as described above). The target
object is then placed at a fixed distance $d{=}1.6$ units along one of the
four cardinal directions (left, right, front, behind) relative to the camera
viewpoint. The question template is: ``Where is the \emph{target} relative to the
\emph{reference}?'' All images are rendered at 336\(\times\)336 resolution and resized
automatically by each model's vision processor during inference.

\begin{figure*}[t]
\centering
\includegraphics[width=\textwidth]{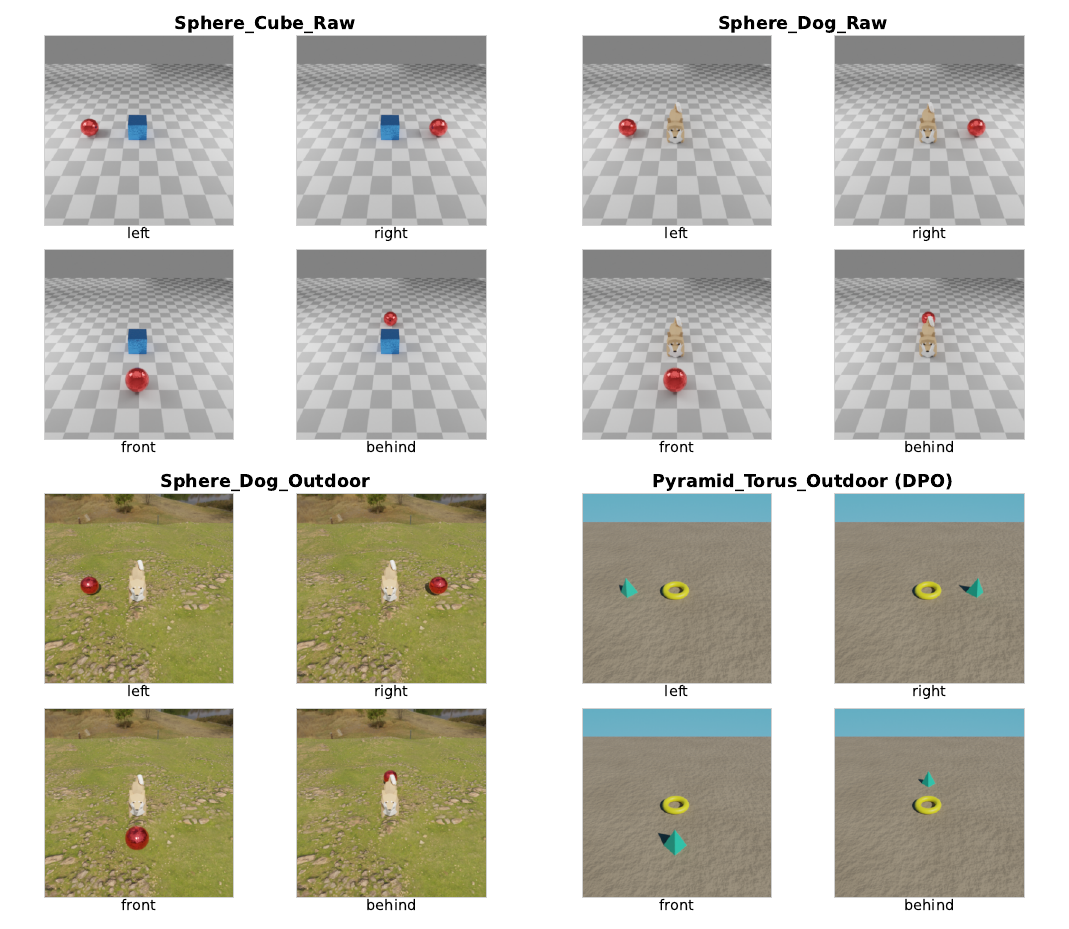}
\caption{\textbf{Representative examples from the controlled synthetic and DPO
training datasets.}
Each panel shows one paired scene from a single dataset. Within a panel, the
object pair, lighting, and camera angle are fixed while the ground-truth
relation cycles through \emph{left}, \emph{right}, \emph{front}, and
\emph{behind}.}
\label{fig:app_dataset_examples}
\end{figure*}

Figure~\ref{fig:app_dataset_examples} shows representative examples from the
three evaluation datasets and the disjoint DPO training dataset: within each
panel, only the spatial relation changes while all other scene elements remain
fixed.

\section{Model Details}
\label{app:model_details}

Table~\ref{tab:models_overview} lists the full set of 9 models evaluated in
this study, along with their vision encoders and LLM backbones.

\begin{table*}[t]
\centering
\small
\setlength{\tabcolsep}{5pt}
\begin{tabular*}{\textwidth}{@{\extracolsep{\fill}}lllll@{}}
\toprule
\textbf{Family} & \textbf{Checkpoint} & \textbf{Vision encoder} & \textbf{LLM backbone} & \textbf{Reference} \\
\midrule
LLaVA & LLaVA-1.5-7B              & CLIP-ViT-L/14   & Vicuna-7B & \citealp{Liu_2024_CVPR} \\
       & LLaVA-v1.6-Vicuna-7B     & CLIP-ViT-L/14   & Vicuna-7B & \citealp{liu2024llavanext} \\
       & LLaVA-v1.6-Mistral-7B    & CLIP-ViT-L/14   & Mistral-7B & \citealp{liu2024llavanext} \\
\addlinespace
Qwen   & Qwen2-VL-2B              & Qwen2-ViT       & Qwen2-1.5B & \citealp{wang2024qwen2vlenhancingvisionlanguagemodels} \\
       & Qwen2-VL-7B              & Qwen2-ViT       & Qwen2-7B & \citealp{wang2024qwen2vlenhancingvisionlanguagemodels} \\
       & Qwen3-VL-2B              & Qwen3-ViT       & Qwen3-1.7B & \citealp{bai2025qwen3vltechnicalreport} \\
       & Qwen3-VL-8B              & Qwen3-ViT       & Qwen3-8B & \citealp{bai2025qwen3vltechnicalreport} \\
\addlinespace
Intern & InternVL2-1B              & InternViT-300M  & Qwen2-0.5B & \citealp{chen2025expandingperformanceboundariesopensource} \\
       & InternVL2.5-1B            & InternViT-300M  & Qwen2.5-0.5B & \citealp{chen2025expandingperformanceboundariesopensource} \\
\bottomrule
\end{tabular*}
\caption{\textbf{Models used in this study.} We evaluate 9 open-weight MLLMs
from three architecture families and four types of vision encoders. Checkpoint
names encode the nominal model scale. InternVL2 has no dedicated paper;
following its model card, we cite the InternVL~2.5 report,
which states that the 2.0 and 2.5 releases share the same architecture.}
\label{tab:models_overview}
\end{table*}

\begin{table*}[t]
\centering
\small
\begin{tabular}{@{}lllr@{}}
\toprule
\textbf{Checkpoint} & \textbf{HF repository} & \textbf{License} & \textbf{Parameters} \\
\midrule
LLaVA-1.5-7B           & \href{https://huggingface.co/llava-hf/llava-1.5-7b-hf}{\nolinkurl{llava-hf/llava-1.5-7b-hf}}           & llama2     & 7.063B \\
LLaVA-v1.6-Vicuna-7B  & \href{https://huggingface.co/llava-hf/llava-v1.6-vicuna-7b-hf}{\nolinkurl{llava-hf/llava-v1.6-vicuna-7b-hf}}  & llama2     & 7.063B \\
LLaVA-v1.6-Mistral-7B & \href{https://huggingface.co/llava-hf/llava-v1.6-mistral-7b-hf}{\nolinkurl{llava-hf/llava-v1.6-mistral-7b-hf}} & apache-2.0 & 7.567B \\
Qwen2-VL-2B           & \href{https://huggingface.co/Qwen/Qwen2-VL-2B-Instruct}{\nolinkurl{Qwen/Qwen2-VL-2B-Instruct}}         & apache-2.0 & 2.209B \\
Qwen2-VL-7B           & \href{https://huggingface.co/Qwen/Qwen2-VL-7B-Instruct}{\nolinkurl{Qwen/Qwen2-VL-7B-Instruct}}         & apache-2.0 & 8.291B \\
Qwen3-VL-2B           & \href{https://huggingface.co/Qwen/Qwen3-VL-2B-Instruct}{\nolinkurl{Qwen/Qwen3-VL-2B-Instruct}}         & apache-2.0 & 2.128B \\
Qwen3-VL-8B           & \href{https://huggingface.co/Qwen/Qwen3-VL-8B-Instruct}{\nolinkurl{Qwen/Qwen3-VL-8B-Instruct}}         & apache-2.0 & 8.767B \\
InternVL2-1B          & \href{https://huggingface.co/OpenGVLab/InternVL2-1B}{\nolinkurl{OpenGVLab/InternVL2-1B}}            & mit        & 0.938B \\
InternVL2.5-1B        & \href{https://huggingface.co/OpenGVLab/InternVL2_5-1B}{\nolinkurl{OpenGVLab/InternVL2_5-1B}}          & mit        & 0.938B \\
\bottomrule
\end{tabular}
\caption{\textbf{Checkpoint repositories and model-card metadata.} Repository
IDs are Hugging Face identifiers. License fields are taken from the official
model cards at the time of writing. Parameter counts are Hugging Face
repository metadata reported to three decimals; model-card pages may display
rounded nominal sizes.}
\label{tab:checkpoint_repos}
\end{table*}

\paragraph{Checkpoint use.}
We use the public checkpoints in Table~\ref{tab:checkpoint_repos} for research
evaluation and LoRA-adapter training/evaluation under the terms listed on their
model cards. We do not redistribute the base checkpoints.

\section{Inference Settings and Answer Extraction}
\label{app:inference_settings}

All reported inference results use the Hugging Face repository identifiers in
Table~\ref{tab:checkpoint_repos}. We load each checkpoint with its
family-specific processor or remote-code interface and use the corresponding
chat/image formatting, while keeping the task instruction and answer-letter
format fixed across model families.

\paragraph{Implementation packages.}
Inference is implemented in PyTorch~\citep{NEURIPS2019_bdbca288} with
Hugging Face Transformers~\citep{wolf-etal-2020-transformers}, loading
weights in bfloat16 with each family's processor (InternVL through its
remote-code interface). LoRA-DPO uses PEFT adapters on the LLM
self-attention and MLP projections, and linear probing uses
scikit-learn~\citep{JMLR:v12:pedregosa11a}.

\paragraph{Decoding.}
Generation is deterministic: sampling is disabled and each prompt is decoded
with at most eight new tokens. The prompt always asks the model to answer with
only the option letter. For each generated response, we first parse a valid
letter answer directly from the stripped response. If the generated text is not
a valid option letter, we fall back to the first-step logits and choose the
highest-scoring valid letter for that prompt's option set.

\section{Prompt Configurations}
\label{app:prompt_configurations}

\begin{figure*}[t]
\centering
\includegraphics[width=\textwidth]{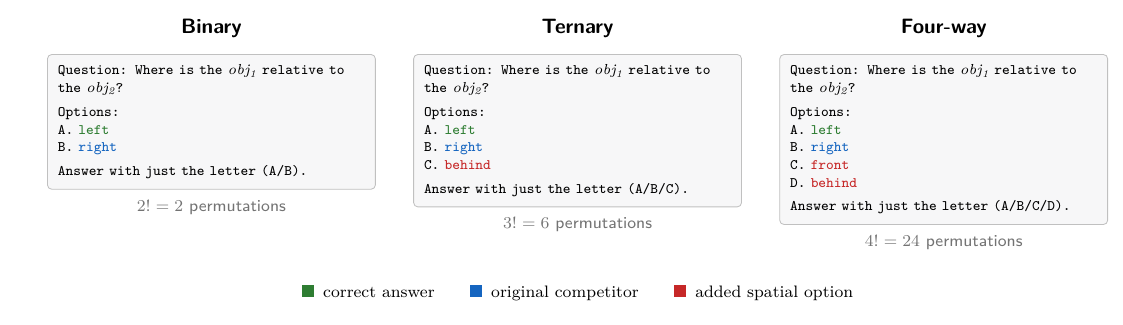}
\caption{\textbf{Prompt formats used in this study.}
\emph{Binary}: two options (correct answer + one competitor).
\emph{Ternary}: a plausible spatial alternative is added.
\emph{Four-way}: all four spatial relations appear as options.
Each configuration is evaluated under all option orderings. The prompt-wise
diagnostic averages over these prompts; the main sample-wise metric counts a
sample as correct only if all orderings are correct.}
\label{fig:prompt_formats}
\end{figure*}

Figure~\ref{fig:prompt_formats} gives the full prompt-format schematic used by
the evaluation. The irrelevant-option control in Section~\ref{sec:irrelevant}
uses the same ternary template, but replaces the added spatial alternative with
an irrelevant option.

\section{Prompt-Wise Permutation-Averaged Evaluation}
\label{app:permutation_eval}

In addition to the sample-wise robust accuracy used in the main case-mining
analysis, we also compute a prompt-wise permutation-averaged diagnostic. A
common practice for reducing position bias is to randomly shuffle the option
order once per question. We use a stronger variant: \emph{exhaustive
shuffling}, where every question is evaluated under \emph{all} \(n!\) possible
orderings and prompt-level accuracy is averaged across them. For a binary
question (\(n{=}2\)) with options \emph{left} and \emph{right}, we run both
orderings and average:
\begin{equation*}
  \text{Acc}_{\text{perm}}
  = \tfrac{1}{2}\!\bigl(\text{Acc}_{\text{(A.left, B.right)}}
  + \text{Acc}_{\text{(A.right, B.left)}}\bigr).
\end{equation*}
For a ternary question (\(n{=}3\)) with options \emph{left}, \emph{right}, and
\emph{behind}, there are \(3!=6\) orderings---covering not only every possible
position for the correct answer, but also every relative arrangement of the two
competing options. The prompt-wise score averages over all six:
\begin{equation*}
  \text{Acc}_{\text{perm-avg}}
  = \tfrac{1}{6}\sum_{k=1}^{6}\text{Acc}_{\text{ordering}\;k}.
\end{equation*}
Because the full permutation set is balanced by construction, every option
appears at every position equally often. This makes prompt-wise
permutation-averaged accuracy useful for diagnosing position effects and for
reporting model--dataset summaries, but binary-stable but
ternary-fragile selection uses the stricter sample-wise score from
Section~\ref{sec:setting}.
Four-way prompts are handled analogously with \(4!=24\) orderings.

\section{Correct-Answer Confidence Analysis}
\label{app:confidence}

Table~\ref{tab:confidence} gives the per-model values behind the
confidence result reported in Section~\ref{sec:highhigh}, with $t$, $c$,
and $a$ denoting the true relation, the binary competitor, and the added
alternative. The unit of analysis is a ternary $a$-capture event: each retained ternary permutation
whose post-fallback response selects the added option $a$ is paired with
the binary $\{t,c\}$ prompt obtained by deleting $a$ while preserving the
relative $t$/$c$ order. A binary record is therefore reused when several
placements of $a$ map to the same binary ordering. Probabilities are
first-step option-letter probabilities renormalized over the displayed
choices, and aggregation proceeds from prompt to case to model
macro-average. The matching covers 67{,}381 prompts, that is, 97.7\% of all
ternary errors on the 55 cases.

\begin{table}[t]
\centering
\small
\setlength{\tabcolsep}{4pt}
\renewcommand{\arraystretch}{1.04}
\setlength{\abovecaptionskip}{3pt}
\begin{tabular}{@{}lcc@{}}
\toprule
\textbf{Model} & $P(t)$ in $\{t,c\}$ & $P(t)$ in $\{t,c,a\}$ \\
\midrule
\rowcolor{gray!8}InternVL2-1B & $0.867 \pm 0.102$ & $0.290 \pm 0.065$ \\
InternVL2.5-1B & $0.841 \pm 0.068$ & $0.282 \pm 0.042$ \\
\rowcolor{gray!8}LLaVA-1.5-7B & $0.770 \pm 0.051$ & $0.294 \pm 0.068$ \\
LLaVA-v1.6-Mistral-7B & $0.719 \pm 0.094$ & $0.348 \pm 0.054$ \\
\rowcolor{gray!8}LLaVA-v1.6-Vicuna-7B & $0.814 \pm 0.094$ & $0.353 \pm 0.064$ \\
Qwen2-VL-2B & $0.792 \pm 0.033$ & $0.387 \pm 0.047$ \\
\rowcolor{gray!8}Qwen3-VL-8B & $0.995 \pm 0.005$ & $0.126 \pm 0.029$ \\
\midrule
\textbf{Average} & $0.828 \pm 0.088$ & $0.297 \pm 0.085$ \\
\bottomrule
\end{tabular}
\caption{\textbf{Correct-answer confidence on added-option errors.}
$P(t)$ is the candidate-normalized first-step probability of the correct
option. The model selects $a$ on ternary prompts; matched binary prompts
remove only $a$, with model, case, sample, and $t$/$c$ order fixed. Entries
are mean $\pm$ SD across case means; Average macro-averages seven models.}
\label{tab:confidence}
\end{table}

\section{Four-Way Persistence of Spatial Lexical Bias}
\label{app:three_to_four_extension}

This appendix supports the four-way extension claim in
Section~\ref{sec:irrelevant}: spatial lexical bias persists when all four
spatial labels are made available. For each selected BSTF case we evaluate
four prompt sets, writing $t$, $c$, and $a$ for the true relation, the binary
competitor, and the added alternative: the original binary $\{t,c\}$ and
ternary $\{t,c,a\}$, plus $\{t,c,o\}$ (replacing the added option by the
remaining fourth relation $o$)
and the full four-way $\{t,c,o,a\}$. Accuracies use sample-wise robust scoring
(all 2, 6, or 24 answer-order permutations correct, respectively). A case is
counted toward the extension if $\mathrm{Acc}(\{t,c\})\ge0.98$,
$\mathrm{Acc}(\{t,c\})-\mathrm{Acc}(\{t,c,a\})\ge0.30$,
$\mathrm{Acc}(\{t,c,o\})\ge0.75$, and
$\mathrm{Acc}(\{t,c,o\})-\mathrm{Acc}(\{t,c,o,a\})\ge0.15$.
We call the cases that pass this test the \emph{strict} subset; the
\emph{strong} subset additionally requires $\mathrm{Acc}(\{t,c,o\})\ge0.85$
and a three-to-four drop of at least $0.30$, and the \emph{very strong}
subset requires $\mathrm{Acc}(\{t,c,o\})\ge0.90$ and a drop of at least
$0.50$.

\begin{table}[t]
\centering
\scriptsize
\setlength{\tabcolsep}{3pt}
\begin{tabular}{lrrrrrr}
\toprule
\textbf{Subset} & \(n\) & \(\{t,c\}\) & \(\{t,c,a\}\) &
\(\{t,c,o\}\) & \(\{t,c,o,a\}\) & \textbf{3\(\to\)4 drop} \\
\midrule
All 55 & 55 & 1.000 & 0.022 & 0.502 & 0.038 & 0.464 \\
Strict & 23 & 1.000 & 0.032 & 0.972 & 0.087 & 0.885 \\
Strong & 19 & 1.000 & 0.017 & 0.988 & 0.024 & 0.964 \\
Very strong & 18 & 1.000 & 0.018 & 0.995 & 0.025 & 0.969 \\
\bottomrule
\end{tabular}
\caption{\textbf{Same-added-option extension from ternary to four-way prompts.}
Entries are macro-averaged per-case sample-wise robust accuracies. \(a\) is the same added
spatial option associated with the original binary-to-ternary drop, and \(o\) is the
remaining fourth spatial relation. The strict, strong, and very-strong subsets
are defined in the text of this appendix.}
\label{tab:three_to_four_extension}
\end{table}

Table~\ref{tab:three_to_four_extension} reports the four accuracies for all
55 cases and for the subsets that pass the test. Under this criterion, 23 of
the 55 selected BSTF cases satisfy the four-way extension test, spanning all three architecture families (10 LLaVA-family
cases, 9 Intern cases, and 4 Qwen cases). The added spatial alternative is
most often \emph{behind} (15/23), followed by \emph{right} (4/23) and
\emph{left} (4/23), and
Table~\ref{tab:app_three_to_four_error_attribution} attributes the erroneous
four-way prompts to the option they select. This extension is a follow-up
diagnostic; the main
behavioral pool remains the 55 BSTF cases defined in
Section~\ref{sec:highhigh}.

\begin{table}[t]
\centering
\small
\setlength{\tabcolsep}{4pt}
\begin{tabular}{llrrr}
\toprule
                & \textbf{Prompt}      &                  & \textbf{Errors}     & \\
\textbf{Subset} & \textbf{set}         & \textbf{Errors}  & \textbf{sel.\ \(a\)} & \textbf{Share} \\
\midrule
All 55 & \(\{t,c,a\}\) & 68,983 & 67,381 & 97.7\% \\
All 55 & \(\{t,c,o,a\}\) & 357,040 & 258,376 & 72.4\% \\
Strict & \(\{t,c,a\}\) & 27,937 & 26,914 & 96.3\% \\
Strict & \(\{t,c,o,a\}\) & 115,436 & 112,348 & 97.3\% \\
\bottomrule
\end{tabular}
\caption{\textbf{Prompt-level error attribution in the four-way extension.} Rows count individual erroneous prompts within the sample-wise extension analysis. The original added option \(a\) is the spatial option associated with the selected binary-stable but ternary-fragile case.}
\label{tab:app_three_to_four_error_attribution}
\end{table}

\paragraph{Same-count replacement control.}
For each of the 55 BSTF cases we also evaluate the two-option prompt
$\{t,a\}$, in which the binary competitor $c$ is replaced by the added
alternative $a$, keeping the option count at two and scoring strictly over
both answer orders. Mean $\mathrm{Acc}(\{t,a\})$ is 0.163 with a median of
0.002; 27 of the 55
cases score exactly 0 and 51 score below 0.90, while only 3 cases remain
robust. Per-model means range from 0.015 for Qwen2-VL-2B to 0.564 for
InternVL2.5-1B. Replacing $c$ by the remaining fourth relation $o$ instead gives a mean
$\mathrm{Acc}(\{t,o\})$ of 0.532, so the collapse is tied to the specific added
word rather than to any change of competitor.

\section{Irrelevant-Option Control Details}
\label{app:irrelevant_details}

Section~\ref{sec:irrelevant} reports sample-wise robust accuracy, where a
sample is correct only if all ternary permutations are answered correctly.
Table~\ref{tab:app_irrelevant_by_model} breaks this sample-wise irrelevant
drop down by model. To inspect how answer position interacts with the
spatial-vs-irrelevant contrast, we additionally compute a prompt-level
position breakdown on the same 55 selected BSTF cases
(Table~\ref{tab:app_irrelevant_position_drop}).

\begin{center}
\centering
\small
\setlength{\tabcolsep}{8pt}
\resizebox{\columnwidth}{!}{%
\begin{tabular}{lrr}
\toprule
\textbf{Model} & \textbf{Cases} & \textbf{Mean irrelevant drop (pp)} \\
\midrule
InternVL2.5-1B & 9 & 37.3 \\
InternVL2-1B & 5 & 33.7 \\
LLaVA-v1.6-Mistral-7B & 11 & 17.4 \\
LLaVA-1.5-7B & 16 & 10.9 \\
LLaVA-v1.6-Vicuna-7B & 9 & 4.2 \\
Qwen3-VL-8B & 3 & 2.8 \\
Qwen2-VL-2B & 2 & 0.0 \\
\bottomrule
\end{tabular}
}
\captionof{table}{\textbf{Per-model irrelevant third-option drop.} Drops use the same sample-wise robust accuracy definition as Section~\ref{sec:irrelevant} and are averaged over the three irrelevant words for the selected BSTF cases available for each model.}
\label{tab:app_irrelevant_by_model}
\end{center}

\begin{center}
\centering
\scriptsize
\setlength{\tabcolsep}{4pt}
\begin{tabular}{lrrrr}
\toprule
\textbf{Added}    & \textbf{Spatial}   & \textbf{Irrelevant} & \textbf{Spatial}    & \textbf{Spatial added-} \\
\textbf{position} & \textbf{drop (pp)} & \textbf{drop (pp)}  & \textbf{excess (pp)} & \textbf{error share} \\
\midrule
A & 72.0 & 4.1 & 68.0 & 99.0\% \\
B & 30.9 & 4.1 & 26.8 & 95.0\% \\
C & 24.3 & 0.9 & 23.4 & 97.3\% \\
\bottomrule
\end{tabular}
\captionof{table}{\textbf{Prompt-level drop by added-option position.} Drops are computed as \(1-\mathrm{prompt\ accuracy}\) within each A/B/C stratum. The spatial-excess column is descriptive: spatial drop minus irrelevant drop under the same added-position stratum.}
\label{tab:app_irrelevant_position_drop}
\end{center}

Together, these diagnostics show that irrelevant third options create
substantial model-dependent difficulty
(Table~\ref{tab:app_irrelevant_by_model}), and that the spatial added-option
lexical-bias pattern is position-amplified relative to the irrelevant
control, especially at the first answer slot
(Table~\ref{tab:app_irrelevant_position_drop}).

\section{Position Bias Breakdown by Dataset and True Label}
\label{app:position_bias_breakdown}

The main text uses position bias only as a diagnostic motivation for
permutation-complete evaluation. Here we give the full prompt-level analysis.
Unlike the main sample-wise robust score, this diagnostic treats each permuted
prompt as the unit of evaluation (as in Appendix~\ref{app:permutation_eval}): we split predictions by the position of the
correct answer and compute accuracy separately for each position. We report
binary, ternary, and a separate fully permuted four-way sweep using \(4!=24\)
orderings per image.

The family-level summary in Table~\ref{tab:family_posbias_summary} shows
distinct sensitivity profiles across the three architecture families.
LLaVA-family models show strong, consistent position bias across all three
formats and always prefer the first answer position; their aggregate $R$
ranges from $0.33$ to $0.67$. Qwen models show the weakest aggregate bias,
with $R\!\ge\!0.85$ in every (format, dataset) cell. The two retained
InternVL checkpoints (InternVL2-1B and InternVL2.5-1B) are near-unbiased in
binary ($R\!\ge\!0.85$) but become moderately position-sensitive in ternary
and four-way prompts ($R$ as low as $0.56$), with the first answer position
remaining preferred on every dataset.

\begin{table}[t]
\centering
\scriptsize
\setlength{\tabcolsep}{2.5pt}
\begin{tabular}{llcccccc}
\toprule
\textbf{Family} & \textbf{Fmt.} & \textbf{WU} & \textbf{SM} & \textbf{VS} & \textbf{CS} & \textbf{DR} & \textbf{DO} \\
\midrule
LLaVA ($n{=}3$)
 & 2-way & 0.63\textsubscript{A} & 0.54\textsubscript{A} & 0.59\textsubscript{A} & 0.47\textsubscript{A} & 0.55\textsubscript{A} & 0.67\textsubscript{A} \\
 & 3-way & 0.57\textsubscript{A} & 0.50\textsubscript{A} & 0.54\textsubscript{A} & 0.48\textsubscript{A} & 0.51\textsubscript{A} & 0.66\textsubscript{A} \\
 & 4-way & 0.58\textsubscript{A} & 0.44\textsubscript{A} & 0.53\textsubscript{A} & 0.40\textsubscript{A} & 0.33\textsubscript{A} & 0.48\textsubscript{A} \\
\addlinespace
Qwen ($n{=}4$)
 & 2-way & 0.99\textsubscript{A} & 0.98\textsubscript{B} & 0.95\textsubscript{B} & 0.95\textsubscript{A} & 1.00\textsubscript{A} & 0.95\textsubscript{B} \\
 & 3-way & 0.96\textsubscript{A} & 0.99\textsubscript{B} & 0.98\textsubscript{B} & 0.90\textsubscript{A} & 0.95\textsubscript{B} & 0.93\textsubscript{B} \\
 & 4-way & 0.92\textsubscript{B} & 0.93\textsubscript{B} & 0.85\textsubscript{B} & 0.90\textsubscript{B} & 0.87\textsubscript{B} & 0.87\textsubscript{B} \\
\addlinespace
Intern ($n{=}2$)
 & 2-way & 0.85\textsubscript{A} & 0.87\textsubscript{A} & 0.97\textsubscript{A} & 0.96\textsubscript{A} & 0.94\textsubscript{B} & 0.91\textsubscript{B} \\
 & 3-way & 0.64\textsubscript{A} & 0.56\textsubscript{A} & 0.58\textsubscript{A} & 0.69\textsubscript{A} & 0.64\textsubscript{A} & 0.74\textsubscript{A} \\
 & 4-way & 0.63\textsubscript{A} & 0.65\textsubscript{A} & 0.58\textsubscript{A} & 0.66\textsubscript{A} & 0.61\textsubscript{A} & 0.81\textsubscript{A} \\
\bottomrule
\end{tabular}
\caption{\textbf{Family-level position bias ratio by dataset.}
Each cell shows $R = \text{Acc}_{\min}/\text{Acc}_{\max}$ after averaging
per-position accuracies over models in the family for that dataset;
subscript indicates the highest-accuracy position.}
\label{tab:family_posbias_summary}
\end{table}

We summarize position bias as a single scalar per condition: the ratio
$R = \text{Acc}_{\min} / \text{Acc}_{\max}$ across option positions.
$R{=}1.0$ means no bias; $R{\to}0$ means severe bias. Unlike the absolute
spread $\Delta = \max - \min$, $R$ is a relative measure that can be compared
across models with different overall accuracy levels. The subscript on each
cell of Table~\ref{tab:family_posbias_summary} indicates which position
achieves the highest accuracy.

\section{Attention Statistics over All BSTF Cases}
\label{app:attention_allcases}

\begin{figure*}[p]
\centering
\includegraphics[width=.92\textwidth]{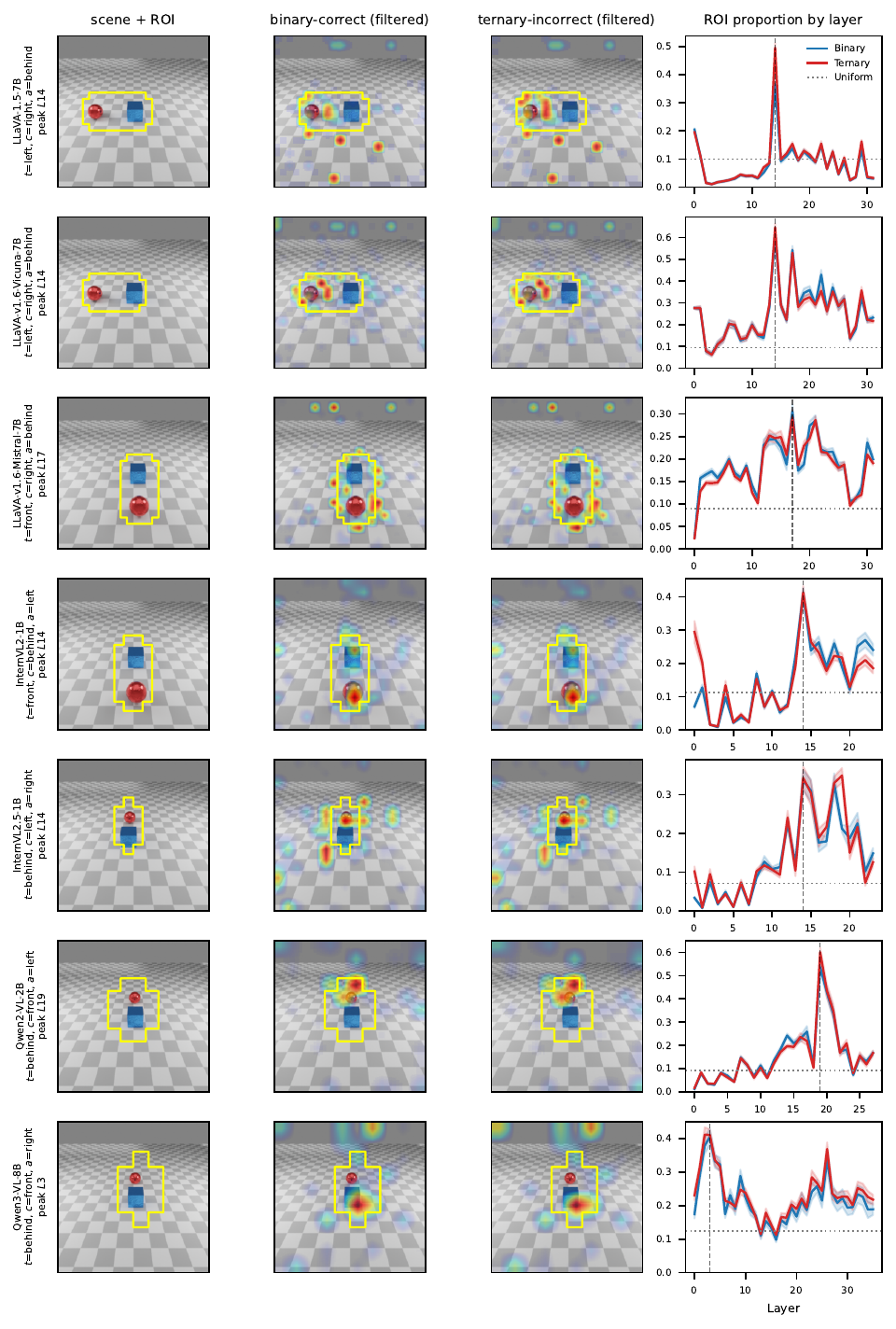}
\caption{\textbf{Attention panels for all seven case-bearing models.}
Each row shows one representative scene for one model. The left column
shows the scene with the two-object ROI as a yellow contour. The middle
columns show the mean binary-correct and ternary-incorrect last-token
attention at the case's ROI-proportion peak layer, rendered as in
Figure~\ref{fig:diagnostic_lasttoken_attn}: the case mean is projected out,
negative values are clipped, the display uses log-percentile scaling, and
the map is overlaid on the scene. The right column shows the filtered
per-layer ROI proportion as mean $\pm$ SEM over the case's records; the
dotted line is the uniform baseline and the dashed vertical line marks the
peak layer.}
\label{fig:app_attention_all_models}
\end{figure*}

This appendix extends the two-case attention visualization of
Section~\ref{sec:last_token_attention_visualization} to all 55 BSTF cases
across the seven case-bearing models. For every case we pair binary-correct
and ternary-incorrect prompts within each scene and measure raw
answer-position attention to the two-object ROI at a fixed representative
layer per model (Table~\ref{tab:app_attention_allcases}, Layer column).
Table~\ref{tab:app_attention_allcases} summarizes the result: ROI share does
not systematically decrease from binary to ternary prompts, and the paired
attention maps remain highly correlated in every model ($\bar{r}=0.978$ over
1{,}072 paired scenes). Figure~\ref{fig:app_attention_all_models} shows one
representative scene per model, rendered with the same recipe as
Figure~\ref{fig:diagnostic_lasttoken_attn}, that is, with the case mean
projected out, negative values clipped, and log-percentile display scaling,
and displayed at each case's peak layer of the filtered ROI-proportion
curve. The statistics in Table~\ref{tab:app_attention_allcases} remain raw
and unfiltered, evaluated at the representative layer.

\begin{table}[t]
\centering
\scriptsize
\setlength{\tabcolsep}{1.8pt}
\renewcommand{\arraystretch}{1.04}
\setlength{\abovecaptionskip}{3pt}
\begin{tabular}{@{}lrrcccc@{}}
\toprule
\textbf{Model} & \textbf{Cases} & \textbf{Scenes} & \textbf{Layer}
 & \textbf{Bin.\ ROI} & \textbf{Tern.\ ROI} & $\bar{r}$ \\
\midrule
\rowcolor{gray!8}InternVL2-1B & 5 & 97 & L10 & 0.084 & 0.081 & 0.997 \\
InternVL2.5-1B & 9 & 166 & L10 & 0.044 & 0.044 & 0.996 \\
\rowcolor{gray!8}LLaVA-1.5-7B & 16 & 320 & L14 & 0.250 & 0.342 & 0.942 \\
LLaVA-v1.6-Mistral-7B & 11 & 216 & L14 & 0.260 & 0.277 & 0.993 \\
\rowcolor{gray!8}LLaVA-v1.6-Vicuna-7B & 9 & 179 & L14 & 0.496 & 0.523 & 0.984 \\
Qwen2-VL-2B & 2 & 36 & L12 & 0.035 & 0.031 & 0.999 \\
\rowcolor{gray!8}Qwen3-VL-8B & 3 & 58 & L15 & 0.101 & 0.096 & 0.999 \\
\midrule
\textbf{All cases} & 55 & 1,072 & --- & --- & --- & 0.978 \\
\bottomrule
\end{tabular}
\caption{\textbf{Quantitative attention statistics over all 55 BSTF cases.}
ROI share is the within-image answer-position attention assigned to the
two-object region of interest at the representative layer (binary-correct
vs.\ ternary-incorrect prompts, raw attention, no filtering), aggregated as
scene means, then case means, then a macro average over cases; $\bar{r}$ is the mean scene-level Pearson correlation between
the paired mean attention grids at the same layer.}
\label{tab:app_attention_allcases}
\end{table}

\section{Probing Design and Full Controls}
\label{app:probing_details}

The probing experiment in Section~\ref{sec:spatial_probe} is deliberately
minimal: a single feature (the last-token residual-stream vector), a single task (read
out the true relation from the failure rows), and one image-content control
(real vs.\ blank). This appendix gives the full design, the per-model
breakdown, and the layer-wise pattern.
Following standard cautions for probing
classifiers~\citep{hewitt-liang-2019-designing,belinkov-2022-probing}, we treat the probe
as a retained-information diagnostic, not a causal-use test: it detects whether
the true relation is linearly present, not whether the base model uses that
feature causally.

\paragraph{Evaluation pool.}
The \emph{evaluation} pool is the 7{,}443 ternary added-spatial-option error
rows from the 55 selected BSTF cases of Section~\ref{sec:highhigh}, i.e.\ rows
where the model selected the added spatial option rather than the true relation.
Each model contributes between 280 and 2{,}220 such error rows (see
Table~\ref{tab:probe_per_model}, column $n_{\mathrm{eval}}$).

\paragraph{Training pool.}
The \emph{training} pool is constructed independently of the case pool. For
each model we sample $4 \times 200 = 800$ binary-correct and ternary-correct
inference rows from the broader inference set, balanced across the four
spatial labels. Training rows are \emph{scene-disjoint} from the evaluation pool: all four
sibling relation images of a rendered scene fall on the same side of the
train/eval boundary, and every model has full four-class support.

\paragraph{Probe and metric.}
For each (model, layer) pair the feature is the LLM last-token residual-stream vector.
Features are standardized and fit with L2-regularized multinomial
LogisticRegression ($C=1$, lbfgs, max\_iter $=2000$). For each layer we record
four-way probe accuracy on the eval pool together with the mean margin
$P(\mathrm{true}) - P(\mathrm{added})$ and the pairwise win rate
$\Pr[P(\mathrm{true}) > P(\mathrm{added})]$. The main paper number is the
peak-layer probe accuracy macro-averaged across models. Chance is $0.25$ for
four-way accuracy and $0$ for the margin.

\paragraph{Blank-image control.}
The blank-image control replaces the input image with a white image of the
same canvas size, preserving prompt structure and the AnyRes
image-token counts~\citep{liu2024llavanext}. We re-run extraction \emph{and} probe training end-to-end on the
blank-image last-token residual-stream vectors, so the blank result is not a transfer of the
real-image probe to blank inputs but an honest re-probing under the same
recipe with the visual content removed.

\paragraph{Per-model results.}
Table~\ref{tab:probe_per_model} reports the per-model peak probe accuracy
and the corresponding margin for both image modes; the same data are plotted
in Figure~\ref{fig:probe_per_model}. The real-image probe accuracy is above $0.985$ for every model.

\begin{table*}[t]
\centering
\small
\setlength{\tabcolsep}{5pt}
\renewcommand{\arraystretch}{1.05}
\begin{tabular}{lrrrrrrrrr}
\toprule
& & & & \multicolumn{3}{c}{\textbf{Real image}} & \multicolumn{3}{c}{\textbf{Blank image}} \\
\cmidrule(lr){5-7} \cmidrule(lr){8-10}
\textbf{Model} & $n_{\mathrm{train}}$ & $n_{\mathrm{eval}}$ & $L$
 & peak $\ell$ & acc. & margin & peak $\ell$ & acc. & margin \\
\midrule
InternVL2-1B          & 800 & 685  & 24 & 10 & 1.000 & $+0.974$ & 0  & 0.108 & $-0.368$ \\
InternVL2.5-1B        & 800 & 1139 & 24 & 14 & 1.000 & $+0.993$ & 1  & 0.207 & $-0.264$ \\
LLaVA-1.5-7B          & 800 & 2220 & 32 & 1  & 0.998 & $+0.958$ & 0  & 0.032 & $-0.748$ \\
LLaVA-v1.6-Mistral-7B & 800 & 1442 & 32 & 13 & 0.985 & $+0.932$ & 0  & 0.061 & $-0.441$ \\
LLaVA-v1.6-Vicuna-7B  & 800 & 1257 & 32 & 0  & 1.000 & $+0.956$ & 16 & 0.165 & $-0.176$ \\
Qwen2-VL-2B           & 800 & 280  & 28 & 12 & 1.000 & $+0.945$ & 6  & 0.207 & $-0.182$ \\
Qwen3-VL-8B           & 800 & 420  & 36 & 0  & 1.000 & $+0.992$ & 6  & 0.179 & $-0.238$ \\
\midrule
Macro                 & --  & --   & -- & -- & \textbf{0.998} & $\mathbf{+0.964}$ & -- & 0.137 & $-0.345$ \\
\bottomrule
\end{tabular}
\caption{\textbf{Per-model probing results.} For each model we train a four-way linear probe (standardized features, L2-regularized logistic regression) on 800 label-balanced binary/ternary-correct rows from broader inference, scene-disjoint from the evaluation set. The probe is then evaluated on the 55-case added-spatial-option error rows for that model (column $n_{\mathrm{eval}}$). Peak probe accuracy is reported with the layer $\ell$ at which it is achieved ($L$ is the model's LLM depth); the corresponding mean margin $P(\mathrm{true}) - P(\mathrm{added})$ on the same layer is shown for reference. Chance accuracy is $0.25$; chance for margin is $0$. The blank-image control replaces the input image with a white image of the same canvas size and re-runs both extraction and training end-to-end.}
\label{tab:probe_per_model}
\end{table*}

\paragraph{Layer dynamics.}
The peak layer varies from $\ell=0$ (essentially the vision-encoder output
projected through one self-attention round) to $\ell=14$ across models, but
in every model the real-image probe accuracy already exceeds $0.92$ at $\ell=0$,
indicating that the true spatial relation reaches the decision token early
and is preserved across the LLM stack. The blank-image curves stay below $0.21$ at every layer, with no
layer-wise rise, ruling out a prompt-template-driven artifact.

\subsection{Randomized-Target (Selectivity) Control}
\label{app:probe_selectivity}

Following the control-task methodology of
\citet{hewitt-liang-2019-designing}, we retrain the identical probe on the
same 800 training rows with permuted labels (five fixed permutations per
model) and evaluate on the same scene-disjoint held-out rows against their
true labels (Table~\ref{tab:app_probe_selectivity}). Randomized-target
probes fit the permuted training pool nearly perfectly (macro 0.991) yet
collapse to the 0.25 chance level on held-out scenes (macro
$0.245 \pm 0.032$), while real-label probes reach 0.998 on the same rows.
Probe capacity and label memorization therefore cannot explain the reported
held-out accuracies; this complements the blank-image control (0.137).

\begin{table}[t]
\centering
\scriptsize
\setlength{\tabcolsep}{3.5pt}
\renewcommand{\arraystretch}{1.04}
\setlength{\abovecaptionskip}{3pt}
\begin{tabular}{@{}lccc@{}}
\toprule
 & \textbf{Held-out} & \multicolumn{2}{c}{\textbf{Randomized-target probe}} \\
\cmidrule(l){3-4}
\textbf{Model} & \textbf{(real probe)} & \textbf{Train fit} & \textbf{Held-out} \\
\midrule
\rowcolor{gray!8}InternVL2-1B & 1.000 & $0.980$ & $0.273 \pm 0.060$ \\
InternVL2.5-1B & 1.000 & $0.959$ & $0.256 \pm 0.114$ \\
\rowcolor{gray!8}LLaVA-1.5-7B & 0.998 & $1.000$ & $0.237 \pm 0.041$ \\
LLaVA-v1.6-Mistral-7B & 0.985 & $1.000$ & $0.181 \pm 0.023$ \\
\rowcolor{gray!8}LLaVA-v1.6-Vicuna-7B & 1.000 & $0.998$ & $0.236 \pm 0.050$ \\
Qwen2-VL-2B & 1.000 & $1.000$ & $0.266 \pm 0.074$ \\
\rowcolor{gray!8}Qwen3-VL-8B & 1.000 & $1.000$ & $0.266 \pm 0.106$ \\
\midrule
\textbf{Macro} & 0.998 & $0.991$ & $0.245 \pm 0.032$ \\
\bottomrule
\end{tabular}
\caption{\textbf{Randomized-target (selectivity) control for the four-way
probe.} Randomized-target probes are trained on the same 800 rows with
permuted labels (5 permutations, mean $\pm$ SD) and evaluated on the same
scene-disjoint held-out rows against their true labels; real-label probes fit
the training pool at 1.000 for every model. High train fit with chance-level
($0.25$) held-out accuracy shows the reported probe numbers cannot come from
probe capacity or label memorization.}
\label{tab:app_probe_selectivity}
\end{table}

\section{Component-Level Ablation Details}
\label{app:component_control}

This appendix documents the design and full per-model results of the
component-level ablation reported in Section~\ref{sec:component_control}.

\subsection{Case Pool, Entry Selection, and Counterfactual Construction}

We use the 55 selected binary-stable but ternary-fragile cases from
Section~\ref{sec:highhigh} (binary sample-wise robust accuracy $=1.0$,
ternary drop $\ge 0.80$). For each case, we select 30 paired entries by
the following filter:

\begin{enumerate}
\item Iterate over (sample index, letter permutation) pairs for which the
spatial-added-option ternary prompt is incorrect with the model predicting the
added spatial option.
\item Require all three irrelevant-word counterfactual ternary prompts
(third option $\in$ \{quartz, violin, lantern\})
to be correct for the same (sample, perm).
\item Retain the first 30 such entries per case in deterministic order.
\end{enumerate}

This selection reduces obvious confounders (image content, letter
ordering, prompt structure, presence of a third option) by holding them
fixed between the spatial-added-option and irrelevant prompts. The 55 cases produce
$1{,}650$ retained entries ($55\times 30$); per-model counts range from
$60$ (Qwen2-VL-2B, 2 cases) to $480$ (LLaVA-1.5-7B, 16 cases).

\subsection{Attribution and Intervention}

\paragraph{Activation extraction.} For each retained entry and for each of
four prompts (one spatial added option, three irrelevant), we record the
post-gate MLP intermediate and the residual stream at the added-option
position (and at the last token, for diagnostics) at every transformer
layer.

\paragraph{Paired delta and top-$K$ ranking.} At the added-option position,
the per-entry delta is
\begin{equation*}
\delta_e = \frac{1}{3}\sum_{w\in\{q,v,l\}} a_e^{w} - a_e^{\text{spatial}},
\end{equation*}
and case-averaged $|\delta|$ ranks neurons / channels. Top-$K$ uses the $K$
largest case-averaged $|\delta|$ indices at each layer; random-$K$ is a
uniform random $K$-subset of the same population at the same layer, with a
fixed seed.

\paragraph{Intervention.} The ablation is applied at the added-option
token position only and zeros the selected indices in the forward pass of
the spatial-added-option prompt. The
$K$-grid covers
\[
\begin{aligned}
K\in\{&1,5,10,20,50,100,200,300,\\
&500,750,1000,1500,3000,4000,\\
&6000,8000,10000\},
\end{aligned}
\]
extended to each model's $d_{\text{model}}$ or $d_{\text{ff}}$. Each shard
records, per case and per (layer, $K$, method),
the model's letter prediction on the spatial-added-option prompt; mean
accuracy is computed across the 30 retained entries per case.

\subsection{Per-Model Summary: Peak Accuracy, Best Layer, and Cliff}

Table~\ref{tab:app_v500_per_model} summarizes the top-$K$ knockout effect
per (model, component) along three axes: the peak case-mean accuracy over
the $(K, L)$ grid, the layer $L^\star$ achieving that peak (with relative
depth), and the cliff layer past which best-$K$ accuracy drops below
peak$/3$ (with relative depth). The Full column counts cases fully repaired
(case-mean accuracy $=1.0$ on the 30 retained entries per case).
Residual ablation reaches peak accuracy $\ge 0.92$ on six of the seven
models ($0.70$ on Qwen3-VL-8B) and fully repairs every case on five of
seven models; neuron ablation attains
intermediate strength, with only LLaVA-1.5-7B below $0.50$ peak. The
residual cliff falls between $26\%$ and $67\%$ relative depth across the
seven models; the MLP-neuron cliff is generally earlier ($4\%$--$37\%$),
and on InternVL2-1B and InternVL2.5-1B the neuron repair peaks at the lowest layer ($L{=}0$) and collapses
immediately.

\suspendLineNumbersIfReview
\begin{table*}[t]
\centering
\small
\setlength{\tabcolsep}{4pt}
\renewcommand{\arraystretch}{1.05}
\begin{tabular}{l c cccc cccc}
\toprule
 & & \multicolumn{4}{c}{\textbf{Residual}} & \multicolumn{4}{c}{\textbf{Neuron}} \\
\cmidrule(lr){3-6} \cmidrule(lr){7-10}
\textbf{Model ($L_{\text{total}}$)} & $N_{\text{case}}$ & peak & $L^\star$ & cliff & Full & peak & $L^\star$ & cliff & Full \\
\midrule
LLaVA-1.5-7B (32)         & 16 & 0.92 & 3 (9.7\%)   & 8 (26\%)  & 13/16 & 0.32 & 7 (22.6\%) & 10 (32\%) & 0/16 \\
LLaVA-v1.6-Vicuna-7B (32) & 9  & 1.00 & 0 (0.0\%)   & 13 (42\%) & 9/9   & 0.74 & 6 (19.4\%) & 11 (35\%) & 3/9 \\
LLaVA-v1.6-Mistral-7B (32) & 11 & 1.00 & 1 (3.2\%)  & 13 (42\%) & 11/11 & 0.53 & 1 (3.2\%)  & 11 (36\%) & 3/11 \\
InternVL2-1B (24)          & 5  & 1.00 & 0 (0.0\%)  & 13 (57\%) & 5/5   & 1.00 & 0 (0.0\%)  & 1 (4\%)   & 5/5 \\
InternVL2.5-1B (24)        & 9  & 0.95 & 5 (21.7\%) & 14 (61\%) & 9/9   & 0.91 & 0 (0.0\%)  & 2 (9\%)   & 7/9 \\
Qwen2-VL-2B (28)           & 2  & 1.00 & 0 (0.0\%)  & 18 (67\%) & 2/2   & 0.97 & 5 (18.5\%) & 10 (37\%) & 1/2 \\
Qwen3-VL-8B (36)           & 3  & 0.70 & 12 (34.3\%) & 19 (54\%) & 0/3  & 0.68 & 8 (22.9\%) & 10 (29\%) & 0/3 \\
\bottomrule
\end{tabular}
\caption{\textbf{Per-model summary of top-$K$ knockout on the 55 selected BSTF cases.}
For each (model, component), \emph{peak} is the maximum case-mean accuracy over the $(K, L)$ grid;
$L^{\star}$ is the layer attaining that peak (relative depth $L^{\star}/(L_{\text{total}}-1)$ in
parentheses); \emph{cliff} is the first layer past $L^{\star}$ where best-$K$ accuracy drops below
peak$/3$ (relative depth in parentheses); \emph{Full} counts cases fully repaired
(case-mean accuracy $=1.0$ on the 30 retained entries per case). Baseline mean accuracy is $0.00$--$0.13$
per model by case-pool construction.}
\label{tab:app_v500_per_model}
\end{table*}

\resumeLineNumbersIfReview

\subsection{Sparsity Thresholds}

Table~\ref{tab:app_v500_K_thresholds} reports, per model and component, the
smallest top-$K$ at the best layer reaching $0.50$ and $0.80$ case-mean
accuracy. Residual is markedly sparser than neuron for every model:
residual reaches $0.50$ at $K\in\{5, 20, 100, 750\}$ (model-dependent),
while neuron requires $K\in[5,6000]$ on models that cross $0.50$; only
LLaVA-1.5-7B never crosses $0.50$ within the tested grid.

\suspendLineNumbersIfReview
\begin{table*}[t]
\centering
\small
\setlength{\tabcolsep}{4pt}
\begin{tabular}{l rrr rrr}
\toprule
 & \multicolumn{3}{c}{\textbf{Residual}} & \multicolumn{3}{c}{\textbf{Neuron}} \\
\cmidrule(lr){2-4} \cmidrule(lr){5-7}
\textbf{Model} & $d_{\text{model}}$ & $K_{50\%}$ & $K_{80\%}$ & $d_{\text{ff}}$ & $K_{50\%}$ & $K_{80\%}$ \\
\midrule
LLaVA-1.5-7B & 4096 & 750 & 1500 & 11008 & -- & -- \\
LLaVA-v1.6-Vicuna-7B & 4096 & 20 & 50 & 11008 & 200 & -- \\
LLaVA-v1.6-Mistral-7B & 4096 & 100 & 500 & 14336 & 6000 & -- \\
InternVL2-1B & 896 & 5 & 100 & 4864 & 20 & 200 \\
InternVL2.5-1B & 896 & 5 & 50 & 4864 & 20 & 500 \\
Qwen2-VL-2B & 1536 & 20 & 200 & 8960 & 5 & 200 \\
Qwen3-VL-8B & 4096 & 100 & -- & 12288 & 3000 & -- \\
\bottomrule
\end{tabular}
\caption{\textbf{Minimum top-$K$ thresholds to reach mean accuracy levels per model.} For each model and
component, the smallest $K$ at which the (best-layer) top-$K$ ablation reaches $0.50$ and $0.80$
case-mean accuracy. $d_{\text{model}}$ is the residual stream width; $d_{\text{ff}}$ is the MLP intermediate
width. Entries marked ``--'' did not reach the corresponding threshold within the tested $K$ grid (max
$=d$).}
\label{tab:app_v500_K_thresholds}
\end{table*}

\resumeLineNumbersIfReview

\subsection{Per-Model Layer Profile}
\label{app:component_layer_profile}

\begin{figure*}[p]
\centering
\includegraphics[width=\textwidth]{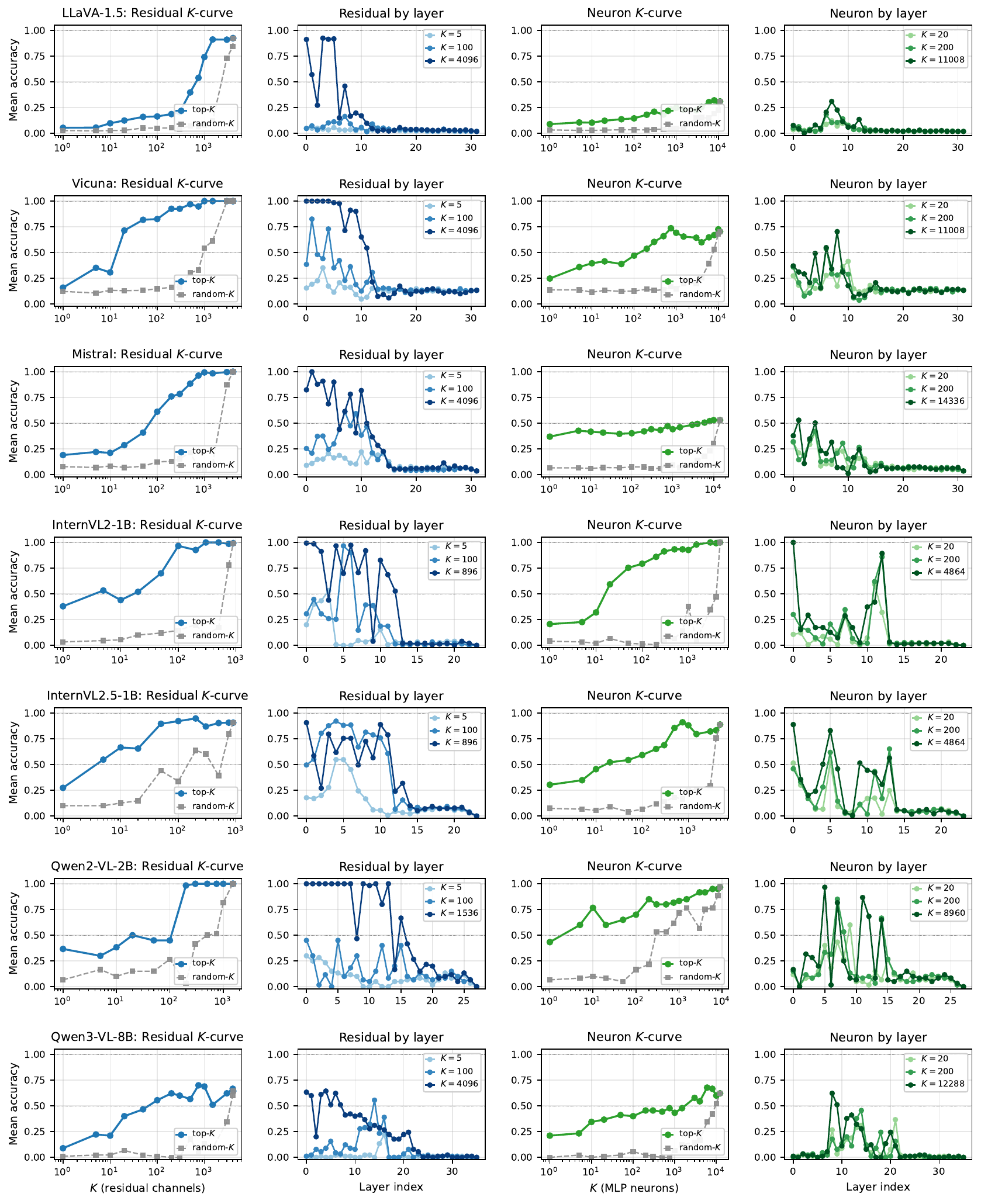}
\caption{\textbf{Per-model knockout result across the seven retained VLMs.}
One row per model; from left to right: residual $K$-curve at the best layer
(top-$K$ in colour, size-matched random-$K$ control in gray dashed),
residual top-$K$ accuracy by layer (three $K$ values), MLP-neuron $K$-curve
(same top-$K$ vs.\ random-$K$ overlay), neuron top-$K$ accuracy by layer
(three $K$ values). For each layer profile, the largest $K$ is the
full-layer setting ($K{=}d_{\text{model}}$ for residual, $K{=}d_{\text{ff}}$
for neuron). The $K$-curve panels confirm that top-$K$ beats random-$K$
across the sparse and mid-range $K$ grid on every model (the two curves
necessarily converge as $K\!\to\!d_{\text{model}}$ or $d_{\text{ff}}$,
since either choice then zeroes the full layer vector); the layer profiles
confirm that the cliff pattern reproduces across all seven models, with
model-dependent cliff depth quantified in
Table~\ref{tab:app_v500_per_model}.}
\label{fig:app_v500_layer_profile}
\end{figure*}

Figure~\ref{fig:app_v500_layer_profile} extends the main-text per-layer view
of Figure~\ref{fig:component_example} to all seven models. For each
(model, component) cell, three top-$K$ values are plotted as accuracy
versus layer, where the largest value is the full-layer setting
($K{=}d_{\text{model}}$ for residual, $K{=}d_{\text{ff}}$ for neuron). The
cliff pattern documented on InternVL2-1B in the main text---repair
collapses past a mid-stack layer beyond which no $K$ rescues the selected
failure---reproduces across all seven models, with model-dependent cliff
depth quantified in Table~\ref{tab:app_v500_per_model}.

\subsection{Top-$K$ vs.\ Random-$K$ Specificity}

Beyond the random-$K$ overlay in Figure~\ref{fig:app_v500_layer_profile},
we tabulated case-averaged top-$K$ vs.\ random-$K$ accuracy at
$K\in\{1,10,100\}$ at each case's best layer. At the single-component level
($K{=}1$), top-$K$ exceeds random-$K$ for all seven models on both
carriers; on the neuron carrier the gap is at least $+0.20$ for six of seven models
(LLaVA-1.5-7B is the smallest, at $+0.10$). The advantage generally grows with $K$ within the sparse regime---at $K{=}100$, neuron
gaps span $+0.15$ to $+0.75$ and residual gaps span $+0.15$ to $+0.88$
across models---and only converges with random-$K$ as $K\to d_{\text{model}}$
or $K\to d_{\text{ff}}$, where either choice zeros the full layer vector at
the added-option position.

\subsection{Sample-Size Caveats}

Per-model case counts are not uniform: LLaVA-1.5-7B has 16 cases,
LLaVA-v1.6-Mistral-7B 11, LLaVA-v1.6-Vicuna-7B 9, InternVL2.5-1B 9,
InternVL2-1B 5, Qwen3-VL-8B 3, Qwen2-VL-2B 2. Qwen2-VL-2B and Qwen3-VL-8B
have very few cases ($n\le 3$), so absolute peak accuracies for those
models are estimated from few-case averages.

\subsection{Process-Level Analysis on InternVL2-1B}
\label{app:process_level}

This analysis studies the interval between the two intervention
endpoints of Sections~\ref{sec:patching} and~\ref{sec:component_control}
under one unified setting. Each failing prompt contains an added spatial
relation word; its correct control replaces that word with an irrelevant
word, with the image, question, option order, and answer roles fixed within
each pair. We apply whole-state patching at either the added-option token or
the last token across all 24 layers of InternVL2-1B. Components and layers were selected on discovery data
only; we report one 30-entry held-out case with aligned answer roles.
Recovery is the fraction of the clean--failing gap in the true-minus-added
logit margin that patching restores (Figure~\ref{fig:process_handoff});
Table~\ref{tab:app_mediation_qkv} gives the sequential mediation test behind
the 63.8\% figure. The L14 last-token value is not directly comparable to
the all-correct plateau of Figure~\ref{fig:patching_summary}, which uses a
binary donor and discrete accuracy rather than a matched ternary donor and
continuous margin recovery.

\begin{table}[t]
\centering
\scriptsize
\setlength{\tabcolsep}{4pt}
\renewcommand{\arraystretch}{1.04}
\setlength{\abovecaptionskip}{3pt}

\begin{tabular}{@{}lcc@{}}
\toprule
\textbf{Sequential mediation (L12 $\to$ L14)} & \textbf{Frac.\ mediated} & \textbf{Source rec.} \\
\midrule
\rowcolor{gray!8}L12 rescue, then reset L14 to failing & 0.638 & 0.817 \\
L12 failing injected, then restore clean L14 & 0.759 & 0.627 \\
\bottomrule
\end{tabular}
\caption{\textbf{Sequential mediation from L12 to L14
(InternVL2-1B, held-out case).} The fraction of the L12 added-option
patching effect removed by resetting the L14 last-token state, and the
reverse test; clamping L14 without changing L12 leaves the baseline
unchanged.}
\label{tab:app_mediation_qkv}
\end{table}

\subsection{Held-Out Functional Validation of Component Sets}
\label{app:component_validation_full}

Sites were fixed before the functional tests (L5 residual channels and
L11 and L12 MLP neurons; Figure~\ref{fig:component_example}B). Within these
layers, components are ranked only on discovery data, frozen, and tested on
the disjoint held-out case against equal-size random sets, changing only the
selected coordinates at the added-option token:
$\text{patched} = \text{failing} + \alpha\,(\text{clean} - \text{failing})$.
Table~\ref{tab:app_component_validation_full} reports the dose grid. At all
three sites selected components outperform random sets, recovery increases
with $\alpha$, and reverse patching (exchanging the clean and failing roles)
moves clean prompts toward failure; the L12 top-500 set accounts for
${\approx}70\%$ of the full L12 MLP effect.

\textbf{Direction-word selectivity.} Using the paper's original
ranking rule (spatial-minus-irrelevant activation delta at the added-option
token at the best repair layer), we profile every model's top-100 residual
channels and top-200 MLP neurons (one-way tests, Bonferroni
$\alpha{=}0.05/K$). All of them separate spatial from irrelevant added words,
and among the six models with more than one observed relation word,
97--100\% of the residual channels and 77--100\% of the MLP neurons further
differentiate \emph{which} relation word was added; Qwen3-VL-8B's case pool
contains a single relation word, so this quantity is not estimable there.
High-ranked components are thus lexically selective at the added-option
token, without implying complete spatial semantics.

\begin{table}[t]
\centering
\scriptsize
\setlength{\tabcolsep}{2.5pt}
\renewcommand{\arraystretch}{1.04}
\setlength{\abovecaptionskip}{3pt}

\begin{tabular}{@{}lcccccc@{}}
\toprule
 & \multicolumn{3}{c}{\textbf{Selected}} & \multicolumn{3}{c}{\textbf{Random}} \\
\cmidrule(lr){2-4}\cmidrule(l){5-7}
\textbf{Intervention} & $\alpha{=}0.5$ & $\alpha{=}1.0$ & $\alpha{=}1.5$ & $\alpha{=}0.5$ & $\alpha{=}1.0$ & $\alpha{=}1.5$ \\
\midrule
\rowcolor{gray!8}L5 res.\ $K{=}200$ & 0.062 & 0.663 & 0.942 & 0.032 & 0.124 & 0.363 \\
L11 MLP $K{=}200$ & 0.111 & 0.190 & 0.237 & 0.014 & 0.018 & 0.022 \\
\rowcolor{gray!8}L12 MLP $K{=}500$ & 0.223 & 0.416 & 0.550 & 0.035 & 0.060 & 0.086 \\
\bottomrule
\end{tabular}
\caption{\textbf{Held-out functional validation: dose
response.} Patched state $=$ failing state $+\,\alpha\,(\text{clean} -
\text{failing})$ on the selected coordinates only; values are normalized
margin recovery on the held-out case. Selected sets respond monotonically in
$\alpha$ and exceed equal-size random sets at every dose; reverse patching
(exchanging clean and failing roles) moves clean prompts toward failure.}
\label{tab:app_component_validation_full}
\end{table}

\subsection{Threshold-Relaxation Sensitivity}
\label{app:threshold_sensitivity}

The strict gate of Section~\ref{sec:highhigh} (binary $=1.00$, drop
$\geq 80$~pp) yields only 2 Qwen2-VL-2B and 3 Qwen3-VL-8B cases, so we
check that the component-level findings do not depend on it. Relaxing the
gate to binary $\geq 0.90$ and drop $\geq 50$~pp expands coverage to 13 and
6 cases, respectively. Defining the cutoff depth as the first layer after
the peak from which the maximum recovery across $K$ stays below one third
of the peak, all four model--component cutoffs are unchanged under the
relaxed set (L16 for Qwen2-VL-2B and L22 for Qwen3-VL-8B, for both
residual-stream and MLP-neuron carriers). Transferring the component sets
ranked on the strict cases to the newly added cases without reselection
also preserves the gap over equally sized matched-random controls (for
example, $0.736 \pm 0.143$ versus $0.155 \pm 0.146$ for the Qwen2-VL-2B
residual stream at layer~1, $K=200$).

\section{DPO Training and Evaluation Details}
\label{app:dpo_training_details}

\subsection{Preference Construction and Training}
\label{app:dpo_training_setup}

\paragraph{Training-pair construction.}
Training uses the tiny disjoint paired synthetic source
Pyramid--Torus--Outdoor, built from a single object pair. Each paired group contains four
images of the same scene, one for each true label
$g\in\{\text{left},\text{right},\text{front},\text{behind}\}$; we split by
paired group with an 80/10/10 train/validation/test split. For a paired group,
choose an option set $S$ of size $k\in\{2,3,4\}$ and an ordered permutation
$\rho(S)$ shown as the answer choices. For every image whose true label
$g\in S$, and for every wrong relation $r\in S\setminus\{g\}$, we create one
preference pair $(x,y^+,y^-)$, where $x$ is the image--prompt input, $y^+$ is
the answer letter assigned to $g$, and $y^-$ is the answer letter assigned to
$r$. Binary and ternary examples enumerate all option sets and all
permutations; four-way examples sample three permutations per paired group.
This yields $24$ binary, $144$ ternary, and $36$ four-way pairs per group, or
$81{,}600$ training pairs and $10{,}200$ validation/test pairs.

The construction is closed-loop within each atomic unit
$(\text{group}, S, \rho)$: for every direction $d$, its count as a chosen
answer equals its count as a rejected answer,
\[
N_u^{+}(d)=N_u^{-}(d)\quad\forall d .
\]
Thus no spatial word or answer letter has a net preference advantage inside the
unit; reducing the loss requires using the image-conditioned relation rather
than a spatial-word or position prior.

\paragraph{DPO objective.}
For DPO, the policy $\pi_\theta$ is the base MLLM with a LoRA adapter enabled;
the reference policy $\pi_{\rm ref}$ is the same model with adapters disabled.
For a pair $(x,y^+,y^-)$, define
\[
\begin{aligned}
\Delta_m &=
\log\pi_m(y^+\mid x)-\log\pi_m(y^-\mid x),\\
\mathcal{L}_{\rm DPO}
&=-\log\sigma\!\left(\beta(\Delta_\theta-\Delta_{\rm ref})\right),
\end{aligned}
\]
where $m\in\{\theta,\mathrm{ref}\}$ and $\beta=0.2$. Both distributions are
read at the next-token answer position after the full image--question prompt.
The chosen and rejected responses are single answer letters, so the loss is
computed only on the corresponding $A$--$D$ token log-probabilities rather than
on free-form text.

\begin{table*}[t]
\centering
\small
\setlength{\tabcolsep}{3pt}
\renewcommand{\arraystretch}{1.04}
\setlength{\abovecaptionskip}{3pt}

\begin{tabular}{@{}lrcccc@{}}
\toprule
\textbf{Model} & \textbf{Cases} & \textbf{Base tern.} & \textbf{SFT tern.} & \textbf{DPO tern.} & \textbf{SFT/DPO bin.} \\
\midrule
\rowcolor{gray!8}InternVL2-1B & 5 & 0.040 & $0.901 \pm 0.058$ & $0.939 \pm 0.014$ & 1.000/1.000 \\
InternVL2.5-1B & 9 & 0.051 & $0.937 \pm 0.032$ & $0.942 \pm 0.004$ & 0.990/0.995 \\
\rowcolor{gray!8}LLaVA-1.5-7B & 16 & 0.006 & $0.836 \pm 0.035$ & $0.826 \pm 0.055$ & 1.000/1.000 \\
LLaVA-v1.6-Mistral-7B & 11 & 0.013 & $0.995 \pm 0.006$ & $0.981 \pm 0.013$ & 1.000/0.999 \\
\rowcolor{gray!8}LLaVA-v1.6-Vicuna-7B & 9 & 0.012 & $0.942 \pm 0.019$ & $0.966 \pm 0.012$ & 1.000/1.000 \\
Qwen2-VL-2B & 2 & 0.050 & $0.997 \pm 0.002$ & $1.000 \pm 0.000$ & 1.000/1.000 \\
\rowcolor{gray!8}Qwen3-VL-8B & 3 & 0.045 & $0.390 \pm 0.054$ & $0.846 \pm 0.011$ & 1.000/1.000 \\
\bottomrule
\end{tabular}
\vspace{5pt}

\begin{tabular}{@{}lp{9.2cm}cc@{}}
\toprule
\textbf{Variant} & \textbf{Added instruction} & \textbf{Tern.\ acc.} & \textbf{Err.\ $\to$ attr.} \\
\midrule
\rowcolor{gray!8}Original & (none) & 2.1\% & 96.5\% \\
Careful inspection & ``Look at the image carefully and consider every option before answering.'' & 8.5\% & 94.4\% \\
\rowcolor{gray!8}Option elimination & ``First eliminate the options that do not match the image, then answer.'' & 4.3\% & 92.5\% \\
Spatial matching & ``Choose the option that best matches the spatial relation in the image.'' & 10.7\% & 95.7\% \\
\bottomrule
\end{tabular}
\caption{\textbf{Repair baselines on the selected BSTF cases.}
Top: SFT versus DPO. SFT trains on the chosen answer letters of the same
81{,}600 preference pairs with cross-entropy loss; both methods update only
LLM LoRA parameters with the same configuration and three seeds (mean
$\pm$ population SD; strict sample-wise robust accuracy). Six of seven
models match DPO within $0.038$. Bottom:
instruction-only prompt variants (base models). Each variant inserts
exactly one sentence before the unchanged final format instruction; strict
sample-wise ternary accuracy over all six orderings on seven models. Even the best variant reaches only 10.7\%, and
errors still overwhelmingly select the added spatial word.}
\label{tab:app_sft_prompt}
\end{table*}

\paragraph{LoRA scope.}
All base parameters are frozen before adapter insertion. LoRA is injected only
into LLM decoder projections: self-attention $q/k/v/o$ and MLP gate/up/down
projections. We use rank $r=8$, $\alpha=16$, dropout $0.05$, and no bias terms.
Only LoRA parameters inside the language model are trainable; the vision
encoder and mm-projector have no trainable parameters.

\paragraph{Optimization.}
For each retained model, we train 3 adapters with different random seeds.
Training uses one epoch, AdamW with learning rate $3{\times}10^{-5}$, zero
weight decay, effective batch size $16$ via gradient accumulation, cosine
learning-rate decay with $5\%$ warmup, and gradient clipping at $1.0$. We
monitor validation KL over answer letters $A$--$D$ relative to $\pi_{\rm ref}$
and per-direction validation accuracy; training stops early if KL exceeds
$1.5$ nats or if any direction drops by more than $5$ points.

\subsection{Repair Evaluation Breakdowns}
\label{app:dpo_repair_breakdowns}

\begin{figure}[H]
\centering
\includegraphics[width=\columnwidth]{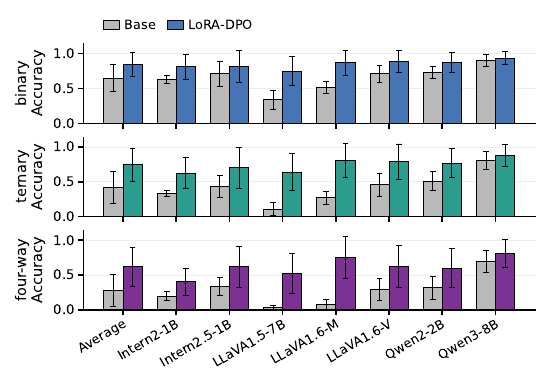}
\caption{\textbf{Strict sample-wise DPO evaluation on the full six-dataset sweep.}
Sample-wise robust accuracy before and after LoRA-DPO.}
\label{fig:app_dpo_samplewise_full6}
\end{figure}

\paragraph{Repair evaluation.}
The main text reports strict sample-wise repair on each adapted model's subset
of the 55 selected BSTF cases. Here we provide the strict sample-wise
six-evaluation-set breakdown for all seven adapted models, averaging the 3 trained
adapters for each model. The first group is the model average; the
remaining groups show per-model means
(Figure~\ref{fig:app_dpo_samplewise_full6}).

\subsection{SFT and Prompting Baselines}
\label{app:sft_prompt}

SFT trains on the chosen answer letter of each of the same 81{,}600
preference pairs with cross-entropy loss; both methods update only the LLM
LoRA parameters, with identical LoRA configuration, three random seeds, and
the vision encoder and projector frozen
(Table~\ref{tab:app_sft_prompt}, top). The prompting baselines
(Table~\ref{tab:app_sft_prompt}, bottom) evaluate the base models on the same
55 cases and all six ternary orderings, inserting exactly one additional
instruction sentence before the
unchanged final format instruction.

\subsection{Transfer to Unseen Relations}
\label{app:unseen_relations}

We evaluate the 206 What'sUp-A images whose gold relation is
\emph{on} or \emph{under}; neither relation appears in repair training. The
two-way condition uses \{\emph{on}, \emph{under}\}; the three-way condition
adds either \emph{left} or \emph{right}; the four-way condition adds both.
Every option order is evaluated on all seven models with the paper's
answer-extraction protocol (Table~\ref{tab:app_unseen_relations}).

\begin{table}[!htb]
\centering
\small
\setlength{\tabcolsep}{4pt}
\renewcommand{\arraystretch}{1.04}
\setlength{\abovecaptionskip}{3pt}

\begin{tabular}{@{}lccc@{}}
\toprule
\textbf{Model / option set} & \textbf{Base} & \textbf{DPO} & $\Delta$ \\
\midrule
\multicolumn{4}{@{}l}{\textit{Four-way (\emph{on}, \emph{under}, \emph{left}, \emph{right}), per model}} \\
\rowcolor{gray!8}LLaVA-1.5-7B & 0.6121 & 0.7150 & $+10.30$~pp \\
LLaVA-v1.6-Vicuna-7B & 0.7053 & 0.8054 & $+10.01$~pp \\
\rowcolor{gray!8}LLaVA-v1.6-Mistral-7B & 0.7282 & 0.8610 & $+13.29$~pp \\
InternVL2-1B & 0.6329 & 0.6337 & $+0.08$~pp \\
\rowcolor{gray!8}InternVL2.5-1B & 0.5643 & 0.6835 & $+11.91$~pp \\
Qwen2-VL-2B & 0.9529 & 0.9864 & $+3.36$~pp \\
\rowcolor{gray!8}Qwen3-VL-8B & 0.9994 & 1.0000 & $+0.06$~pp \\
\midrule
\multicolumn{4}{@{}l}{\textit{Macro over the seven models, by option set}} \\
\rowcolor{gray!8}Two-way (\emph{on}, \emph{under}) & 0.7868 & 0.7972 & $+1.04$~pp \\
Three-way ($+$ \emph{left} or \emph{right}) & 0.7700 & 0.8196 & $+4.96$~pp \\
\rowcolor{gray!8}Four-way ($+$ \emph{left} and \emph{right}) & 0.7421 & 0.8122 & $+7.00$~pp \\
\bottomrule
\end{tabular}
\caption{\textbf{Unseen-relation transfer on What'sUp-A
\emph{on}/\emph{under}.} 206 images whose gold relation (\emph{on} or
\emph{under}) never appears in repair training; every option order is
evaluated on all seven models with the paper's answer-extraction protocol.
Top: per-model four-way accuracy; no model declines
after repair, and the near-zero Qwen3-VL-8B gain reflects its near-ceiling
base accuracy. Bottom: macro average as the option set expands from two to
four options; base accuracy declines while the repair gain grows.}
\label{tab:app_unseen_relations}
\end{table}

\subsection{General-Capability Checks}
\label{app:capability_checks}

Table~\ref{tab:app_capability} reports the change from base to
repaired checkpoints on five benchmarks outside spatial MCQ. MME uses the
full 2{,}374-question test split with official scoring and three DPO seeds
per model; MMBench (dev-EN, circular accuracy over 1{,}164 groups), MMLU
(57 subjects $\times$ 25 questions, zero-shot), IFEval (strict prompt-level
accuracy, 541 prompts), and COCO captioning (CIDEr~\citep{7299087}, 2{,}000-image subset)
average three seeds for LLaVA-1.5-7B and InternVL2.5-1B. Changes are small and two-sided; IFEval and
COCO require free-form generation, so these checks also argue against broad
over-alignment to the answer-letter format.

\begin{table}[!htb]
\centering
\scriptsize
\setlength{\tabcolsep}{2pt}
\renewcommand{\arraystretch}{1.04}
\setlength{\abovecaptionskip}{3pt}

\begin{tabular}{@{}lccccc@{}}
\toprule
\textbf{Model} & \textbf{MME} & \textbf{MMBench} & \textbf{MMLU} & \textbf{IFEval} & \textbf{COCO} \\
 & $\Delta$pp & $\Delta$pp & $\Delta$pp & $\Delta$pp & $\Delta$CIDEr \\
\midrule
\rowcolor{gray!8}InternVL2-1B & $-0.22$ & $+0.25$ & $+0.84$ & $+0.19$ & $+0.0368$ \\
InternVL2.5-1B & $-0.84$ & $-1.18$ & $-0.47$ & $+0.37$ & $+0.0086$ \\
\rowcolor{gray!8}LLaVA-1.5-7B & $-0.11$ & $-0.08$ & $+0.73$ & $+2.34$ & $-0.0126$ \\
LLaVA-v1.6-Mistral-7B & $-0.27$ & $+0.94$ & $+0.21$ & $+2.77$ & $-0.0822$ \\
\rowcolor{gray!8}LLaVA-v1.6-Vicuna-7B & $-2.39$ & $+1.55$ & $+0.14$ & $+0.19$ & $-0.0289$ \\
Qwen2-VL-2B & $+0.51$ & $+0.51$ & $+0.21$ & $+2.96$ & $+0.0080$ \\
\rowcolor{gray!8}Qwen3-VL-8B & $+0.08$ & $+0.00$ & $-0.07$ & $+1.11$ & $+0.0192$ \\
\midrule
\textbf{Macro} & $-0.46$ & $+0.33$ & $+0.19$ & $+1.40$ & $-0.0075$ \\
\bottomrule
\end{tabular}
\caption{\textbf{General-capability checks on the repaired
models.} Change (DPO $-$ base) on MME (full 2{,}374-question test, official
scoring), MMBench dev-EN circular accuracy (1{,}164
groups), MMLU (57 subjects $\times$ 25 questions, zero-shot), IFEval
(strict prompt-level accuracy, 541 prompts), and COCO captioning CIDEr
(2{,}000-image subset); MME averages three seeds per model, and the other
benchmarks average three seeds for LLaVA-1.5-7B and InternVL2.5-1B.}
\label{tab:app_capability}
\end{table}

\section{Option-Free (Open-Ended) Generation}
\label{app:open_ended}

To examine relation-word preferences without any displayed
candidates, we evaluate option-free one-word generation on the three
LLaVA-family models over the complete grid of three controlled synthetic
sets and all four relations (36 settings, 500 images each, 18{,}000
generations). The prompt keeps the original question, removes all choices
and relation hints, and appends ``Answer the question with one word.''
(greedy decoding). Scoring is normalized exact match; because \emph{front}
is conventionally expressed as a multiword phrase in English, we credit unambiguous ``(in) front (of \dots)'' responses as \emph{front}.
Table~\ref{tab:app_oneword} shows the base-model grid: many settings
recover the correct relation without candidates, but accuracy depends
strongly on the target relation, and LLaVA-v1.6-Mistral answers
\emph{right} for all 1{,}500 \emph{front} images. We do not claim that
these generation preferences share the same mechanism as the MCQ effect.

\begin{table}[H]
\centering
\small
\setlength{\tabcolsep}{4pt}
\renewcommand{\arraystretch}{1.04}
\setlength{\abovecaptionskip}{3pt}

\begin{tabular}{@{}llcccc@{}}
\toprule
\textbf{Model} & \textbf{Data} & \emph{left} & \emph{right} & \emph{front} & \emph{behind} \\
\midrule
\rowcolor{gray!8}LLaVA-1.5 & CS & 100\% & 100\% & 0.0\% & 0.0\% \\
\rowcolor{gray!8} & DO & 100\% & 100\% & 74.2\% & 83.8\% \\
\rowcolor{gray!8} & DR & 100\% & 100\% & 0.0\% & 76.0\% \\
LLaVA-1.6-M & CS & 100\% & 100\% & 0.0\% & 0.0\% \\
 & DO & 100\% & 100\% & 0.0\% & 89.0\% \\
 & DR & 100\% & 100\% & 0.0\% & 99.4\% \\
\rowcolor{gray!8}LLaVA-1.6-V & CS & 100\% & 100\% & 0.0\% & 17.2\% \\
\rowcolor{gray!8} & DO & 68.8\% & 99.8\% & 97.0\% & 64.2\% \\
\rowcolor{gray!8} & DR & 90.2\% & 100\% & 100\% & 54.0\% \\
\bottomrule
\end{tabular}
\caption{\textbf{Option-free one-word generation.} Base-model
relation accuracy on the full 36-setting grid. Prompts keep the original
question, remove all candidates and relation hints, and request one word
(``Answer the question with one word.''); greedy decoding, 500 images per
cell. Scoring is normalized exact match, crediting unambiguous ``(in)
front'' responses as \emph{front}.}
\label{tab:app_oneword}
\end{table}

\section{Spatial Fine-Tuning Baseline}
\label{app:newer_models}

As a preliminary comparison on a shared base model, we evaluate the three
strict BSTF cases of Qwen3-VL-8B. The base model reaches 1.000/0.045
binary/ternary accuracy; SenseNova-SI-1.1-Qwen3-VL-8B~\citep{Cai_2026_CVPR}, a spatially
fine-tuned derivative of the same base model, reaches 0.637/0.006; and our
LLM-only DPO repair reaches 1.000/0.846. On these cases, spatial fine-tuning
alone does not necessarily resolve the failure mode. Because only three
cases are available and the training setups differ, this comparison is not a
general ranking of models or training methods.

\end{document}